\documentclass[runningheads]{llncs}
\usepackage[T1]{fontenc}
\usepackage{newtxtext}
\usepackage[varvw]{newtxmath}
\usepackage{amsmath}
\usepackage{xcolor}
\allowdisplaybreaks[2]
\usepackage{booktabs}
\usepackage{algorithm,algorithmicx}
\usepackage{algpseudocode}
\usepackage{graphicx}
\usepackage{afterpage}
\usepackage{textcomp}
\usepackage{multirow}
\usepackage[caption=false,font=footnotesize]{subfig}
\usepackage{cleveref}
\providecommand{\texorpdfstring}[2]{#1}

\newcommand{\backtab}[1]{\hspace{0.6cm}}
\newcommand{\backspace}[1]{\hspace{0.2cm}}
\spnewtheorem{assumption}{Assumption}{\bfseries}{\itshape}
\newcommand{\rrr}[1]{#1}
\DeclareMathOperator{\Tr}{Tr}
\crefname{equation}{}{}
\Crefname{equation}{Equation}{Equations}
\Crefname{assumption}{Assumption}{Assumptions}
\title{A Historical Trajectory Assisted Optimization Method for Zeroth-Order Federated Learning}
\titlerunning{Historical Trajectory Assisted Zeroth-Order Federated Learning}
\author{Yu Cai\inst{1} \and Canxi Huang\inst{1} \and Zike Li\inst{1} \and Jing Gong\inst{2} \and Xiaoyu He\inst{1}\thanks{Corresponding author}}
\authorrunning{Y. Cai et al.}
\institute{School of Software Engineering, Sun Yat-sen University, Zhuhai 519082, China\\
\email{\{caiy86, huangcx23, lizk8\}@mail2.sysu.edu.cn}\\
\email{hexy73@mail.sysu.edu.cn}
\and
School of Computing and Information Systems, The University of Melbourne, Melbourne, VIC, Australia\\
\email{gongjing13424@gmail.com}}

\begin{document}
\maketitle
\begin{abstract}
Federated learning heavily relies on distributed gradient descent techniques. In the situation where gradient information is not available, the gradients need to be estimated from zeroth-order information, which typically involves computing finite-differences along isotropic random directions. This method suffers from high estimation errors, as the geometric features of the objective landscape may be overlooked during the isotropic sampling. 
In this work, we propose a non-isotropic sampling method to improve zeroth-order gradient estimation. The method uses historical solution trajectories to identify a promising subspace and constructs a non-isotropic Gaussian smoothing that balances exploitation in this subspace with global exploration.
We implement this method in zeroth-order federated settings, and show that the convergence rate aligns with existing ones while introducing no significant overheads in communication or local computation. The effectiveness of our proposal is verified on several numerical experiments in comparison to several commonly-used zeroth-order federated optimization algorithms.
The experiments cover both independent and identically distributed data and heterogeneous client data distributions.

\end{abstract}

\section{Introduction}
This work considers solving the unconstrained federated optimization problem~\cite{fang2022communication} defined as
\begin{equation} \label{eq:problem-definition}
    \min_{w \in \mathbb{R}^n} f(w) = \frac{1}{N} \sum_{i=1}^N \mathbb{E}_{\xi \sim \mathcal{D}_i}[F(w;\xi)]
\end{equation}
where $N$ is the number of clients, $\xi$ is a data sample, $\mathcal{D}_i$ is the local data distribution associated with client $i$, and $w \in \mathbb{R}^n$ is the decision vector. The task is to optimize the global objective $f$ with clients only having access to their local data samples. The data distributions $\mathcal{D}_i$ and $\mathcal{D}_j$ may differ for $i \neq j$, which is also known as the non independent and identically distributed (non-IID) setting.

Efficiently solving problem \cref{eq:problem-definition} is the key to implementing federated learning systems. One representative of federated optimization methods is FedAvg~\cite{mcmahan2017communication}. The clients in FedAvg fetch the global parameters from the server, update them using the stochastic gradient descent (SGD) method with their local data, and upload the parameters back to the server for aggregation. The server aggregates the obtained parameters without accessing client data or performing local model updates, which ensures both data privacy and communication efficiency. Various FedAvg variants have been proposed, improving FedAvg in local update~\cite{ma2022communication}, client selection~\cite{luo2022tackling}, and the aggregation mechanism~\cite{zhang2023fedala,shi2024sam}. One of the most common features shared by these methods is that they rely on gradient information for guiding the clients' local updates, measuring the learning progress, or encoding the model to be transferred between the server and clients. The clients thus must support automatic differentiation or have manually coded gradients.

We focus on solving the problem \cref{eq:problem-definition} in zeroth-order settings. That is, the individual objective $F$ is a black box, and for a given vector $x$ and a data sample $\xi$, one can obtain no information other than the objective value $F(x;\xi)$. This simulates the real-world scenario where the structure of the objective is unknown or evaluating the gradients is intractable, typically due to security or commercial considerations. An example that has emerged recently is the federated learning system based on closed-source large language models where the neural architectures are not released publicly. Client-side users in this case cannot evaluate the gradients, making it intractable to deploy downstream tasks such as prompt tuning~\cite{che2023federated,lin2023efficient,sun2023fedbpt,zhao2023fedprompt}. Zeroth-order optimization also reduces memory usage as no backward pass is required~\cite{zhang2024revisiting}; this is particularly useful in certain federated systems where clients involve low-performing devices.

Gaussian smoothing~\cite{gao2022GS} is one of the most popular techniques for zeroth-order optimization, which samples a random vector from an isotropic Gaussian distribution, and outputs the finite-difference along this vector as a gradient estimation.
It is easy to implement, enjoys strong theoretical guarantees, and can be used in combination with established gradient-based algorithms.
However, the estimation output by Gaussian smoothing is biased and noisy, and how accurate the estimation is depends heavily on the objective landscape features~\cite{he2022adaptive}.
This may prevent the convergence of Gaussian smoothing based zeroth-order optimization, as the used isotropic Gaussian cannot fully exploit the objective landscape.

In this paper, we extend Gaussian smoothing to federated optimization using a non-isotropic Gaussian whose covariance is a convex combination of a trajectory-induced low-rank projection and the identity matrix. The two components promote exploitation in a promising subspace and preserve global exploration, respectively. Periodic covariance updates keep the communication overhead negligible, and simulations demonstrate the method's robustness and efficiency.

\paragraph{Notations}
$\mathcal{N}(\mu, C)$ represents a Gaussian distribution with expectation $\mu$ and covariance matrix $C$. $I_n$ denotes an $n\times n$ identity matrix. $\|x\|_2$ denotes the $\ell_2$ norm of a vector $x$. 
$\mathbb{E}[x]$ is the expectation of a random variable $x$. $\Tr[M]$ is the trace of a matrix $M$.

\section{Non-isotropic Gaussian smoothing}
\label{sec:nonisotropic-gaussian-smoothing}
In this section, we assume the decision space $\mathbb{R}^n$ is equipped with some generic vector norm $\|\cdot\|$. Its dual norm is denoted by $\|\cdot\|_*$.

\subsection{Preliminaries on Gaussian smoothing}
We firstly define the Gaussian smoothing based gradient estimation method that involves a non-isotropic covariance matrix. The method relies on the smoothness property of a function, which we detail as below:
\begin{definition}[Smoothness] 
    \label{smoothness}
    We say a function $h$ is $L$-smooth if it is differentiable and its gradient satisfies:
    \begin{equation} 
    \|\nabla h(x) - \nabla h(y)\|_* \leq L\|x-y\|,  \forall x, y 
    \end{equation}
    for some constant $L \in \mathbb{R}_+$.
\end{definition}

The smoothness of a function $h$ implies that it has a quadratic bound:
\begin{equation} \label{func_bound}
|h(y) - h(x) - \langle \nabla h(x),y-x \rangle| \leq \frac{L}{2}\|y-x\|^2, \forall x,y .
\end{equation}

Gaussian smoothing is a method to generate a smooth surrogate for a function $h$:
\begin{definition}[Non-isotropic Gaussian Smoothing] \label{nga}
Given a function $h: \mathbb{R}^n \to \mathbb{R}$, its Gaussian smoothing is defined as:
\begin{equation} \label{nga-equation}
    h_{\mu, C}(x) = \mathbb{E}_{v \sim \mathcal{N}(0,  C)}[h(x+\mu v)]
\end{equation}
where $\mu \in \mathbb{R}_+$ is the smoothing radius and $C \in \mathbb{R}^{n \times n}$ is a positive-definite covariance matrix. The choice $C=I_n$ recovers standard (isotropic) Gaussian smoothing~\cite{ghadimi2013stochastic}; otherwise, the smoothing is non-isotropic. 
\end{definition}

The Gaussian smoothing $h_{\mu, C}$ is differentiable, and its gradient has a closed-form expression that only depends on zeroth-order information~\cite{he2022adaptive}:

\begin{lemma}[Differentiability] \label{lemma:differentiability}
For a function $h: \mathbb{R}^n \to \mathbb{R}$, its Gaussian smoothing is differentiable and has the gradient defined as: 
\begin{equation}
\label{eq:gradient_gaussian_smoothing}
\begin{split}
    \nabla h_{\mu, C}(x)& = \mathbb{E}_{v \sim \mathcal{N}(0, C)}\left[\frac{h(x+\mu v)-h(x-\mu v)}{2\mu}C^{-1}v\right].
\end{split}
\end{equation} 
\end{lemma}

The above shows that we can compute an ascent direction of $h_{\mu,C}$ as 
\begin{equation} \label{eq:unbiased-gradient-estimation-of-gaussian-smoothing}
    g = \frac{h(x+\mu v) - h(x-\mu v)}{2\mu}v
\end{equation}
where $v$ is drawn from $\mathcal{N}(0, C)$. 
It is easy to see
\[
   \mathbb{E}[\langle g, \nabla h_{\mu,C}(x)\rangle] = \|\sqrt{C} \nabla h_{\mu,C}(x)\|^2_2 \ge 0.
\] 
Thus, $g$ is indeed a stochastic ascent direction of $h_{\mu,C}$. Hereinafter we will call $g$ the stochastic gradient of $\nabla h_{\mu,C}(x)$.

The importance of Gaussian smoothing is that it acts as a surrogate to the original function provided the latter is smooth: 

\begin{lemma}[Properties of non-isotropic Gaussian smoothing]
\label{similarity}

Assume a function $h: \mathbb{R}^n \to \mathbb{R}$ is $L$-smooth. Then, the Gaussian smoothing $h_{\mu,C}$  is also $L$-smooth. In addition, its difference to the original function can be bounded as
\begin{equation}\label{eq:gaussian-smoothing-bound-on-gradient}
    \|\nabla h_{\mu,C}(x) - \nabla h(x)\|_* \le  \frac{L\mu}{2} \mathbb{E}[\|v\|^2\|C^{-1} v\|_*],
\end{equation}
and
\begin{equation}\label{eq:gaussian-smoothing-bound-on-objective}
    |h_{\mu,C}(x) - h(x)| \le  \frac{L\mu^2}{2}\mathbb{E}[\|v\|^2],
\end{equation}
where the expectation is taken over $v \sim \mathcal{N}(0,C)$.
\end{lemma}

\subsection{Covariance Matrix Specification}
\label{ss:covariance-matrix-specification}
Lemmas 1 and 2 state that the Gaussian smoothing $h_{\mu,C}$ serves as a good surrogate to the original objective function $h$, as it is smooth and has closed-form gradients.
Optimizing $h_{\mu,C}$ instead of $h$ would nevertheless introduce an approximation error due to the difference between these two functions.
On the other hand, \cref{eq:gaussian-smoothing-bound-on-objective} shows that this error is bounded, and the bound depends on the covariance matrix $C$.
It implies that we can control the covariance matrix to reduce the error, which is exactly why we need the non-isotropic Gaussian smoothing. A trivial choice to minimize the bound in \cref{eq:gaussian-smoothing-bound-on-objective} is letting $C \to 0$.
This is however meaningless, as it would bring rounding errors in practice.

In this work we suggest constraining the covariance matrix as
\begin{equation}\label{eq:covariance-matrix-model}
C = (1-\alpha) I_n + \alpha Q Q^\top,
\end{equation}
where $\alpha \in [0,1)$ is a constant and $Q \in \mathbb{R}^{n\times m}$ is an orthonormal basis matrix satisfying $Q^\top Q = I_m$. Consequently, $Q Q^\top$ is the corresponding projection matrix.
The term $m$ here is a user specified hyperparameter that should be smaller than $n$. The matrix $Q$ is kept orthogonal, as its magnitude is not important: the magnitude of $C$ can be absorbed into the constant $\mu$. With $\alpha \to 1$, the gradients output by Gaussian smoothing are located in a subspace spanned by the columns of $Q$. If these gradients are used for guiding the search, the algorithm will keep staying in this subspace. On the contrary, taking $\alpha \to 0$ leads to the standard Gaussian smoothing, which estimates the gradient in the whole space $\mathbb{R}^n$ and hence encourages the algorithm to explore unexplored regions. In other words, the covariance matrix $C$ controls the balance between exploration and exploitation.

We propose that an appropriate covariance matrix $C$ should capture insensitive directions over the objective landscape. A direction is called sensitive if moving along this direction causes a rapid change on the objective value and vice versa. To see why the sensitiveness matters, consider an ideal setting where the objective is a quadratic function $f(x) = \frac{1}{2}x^\top H x$ with symmetric positive definite Hessian $H\in \mathbb{R}^{n\times n}$.
The objective function is $1$-smooth w.r.t. the Euclidean norm $\|x\|^2 = x^\top H x$ and its dual norm $\|x\|_*^2 = x^\top H^{-1} x$.
In this setting, the matrix $Q$ for minimizing the bound in \cref{eq:gaussian-smoothing-bound-on-objective} can be found by solving the following problem:

\begin{equation}\label{eq:find-the-Q}
    \begin{cases}
    \min \limits_{Q\in \mathbb{R}^{n\times m} } & \mathbb{E}_{v \sim \mathcal{N}(0,C) }[\|v\|^2] \\
    \text{s.t. } & C = (1-\alpha)I_n + \alpha QQ^\top\\
    &Q^\top Q = I_m 
    \end{cases}.
\end{equation}
Since $\mathbb{E}_{v\sim\mathcal{N}(0,C)}[v^\top H v]=\operatorname{tr}(HC)$, problem~\eqref{eq:find-the-Q} is equivalent, up to the constant $(1-\alpha)\operatorname{tr}(H)$, to minimizing $\operatorname{tr}(Q^\top H Q)$ subject to $Q^\top Q=I_m$. By the Rayleigh--Ritz theorem, the minimizer is formed by the eigenvectors of $H$ associated with its $m$ smallest eigenvalues, or equivalently, the $m$ principal eigenvectors of $H^{-1}$. Thus, the optimal $Q$ corresponds to the $m$ most insensitive directions.

\afterpage{
\begin{table}[t]
    \centering
    \resizebox{0.92\columnwidth}{!}{
    \begin{tabular}{lllllllll}
       \toprule
       $\epsilon$ & $10^{-2}$ & $10^{-1}$ & $10^0$ & $10^1$ & $10^2$ & $10^3$ & $10^4$ & $10^5$ \\
       $\gamma(\epsilon)$ & 0.9993 & 0.9423 & 0.8268 & 0.7096 & 0.5931 & 0.4979 & 0.4924 & 0.4919\\
       \bottomrule
    \end{tabular}
    }
    \vspace{0.2cm}\caption{Ratio of correctly capturing insensitive directions by tracking solution trajectories}
    \label{table:ratio-of-correctly-capturing-insensitive-directions}
\end{table}
}

When solving a problem whose Hessian information is not accessible, we can approximate its insensitive directions by tracking the solution trajectories. An intuitive reason is: the algorithm has to repeatedly exploit the subspace spanned by the insensitive directions, as the progress in this subspace is harder to achieve than in that spanned by sensitive directions.
Here we use simulations on a 2-dimensional example to support this statement.
Let us define $f(x) = 10^3 x_1^2 + x_2^2$. So the most sensitive direction is $e^{[1]} = (1,0)^\top$ and the most insensitive one is $e^{[2]} = (0,1)^\top$. 
We consider the stochastic gradient descent iterations $x^{[t+1]} = x^{[t]} - 10^{-4} g^{[t]}$, where gradients are sampled from $g^{[t]} \sim \mathcal{N}(\nabla f(x^{[t]}), \epsilon^2 I_2)$ with $\epsilon$ measuring the noise.
Defining
\[
\gamma(\epsilon) = \frac{1}{T} \sum_{t=0}^{T-1} \mathbb{I}\{|g_2^{[t]}| > |g_1^{[t]}|\}
\]
as the metric for indicating whether the solution trajectories can capture the insensitive direction. 
Here, $|g_2^{[t]}|>|g_1^{[t]}|$ means that the step at iteration $t$ forms an acute angle with the most insensitive direction $e^{[2]}$; hence, a larger $\gamma(\epsilon)$ indicates more frequent recovery of this direction. We run 10 independent simulations for each $\epsilon \in \{10^{-2},10^{-1},\dots,10^5\}$, with $T=10^4$ and $x^{[0]}=(10^3,10^3)^\top$. As shown in \Cref{table:ratio-of-correctly-capturing-insensitive-directions}, the trajectories align with $e^{[2]}$ with high probability at low noise levels.
This experiment states that trajectories of the solutions produced in a search procedure do capture the insensitive directions. Conversely, the matrix $Q$ can be specified in \cref{eq:covariance-matrix-model} with principal components of solution trajectories as they are expected to coincide with the insensitive directions and therefore improve the gradient estimation accuracy. We implement this idea in federated settings in the next section.

\section{The Proposed Method}
\label{sec:implementations}
We apply the non-isotropic Gaussian smoothing described above to the federated optimization problem \cref{eq:problem-definition} and propose the \underline{z}eroth-\underline{o}rder \underline{fed}erated optimization algorithm assisted by \underline{h}istorical \underline{t}rajectories (ZOFedHT).

\subsection{Implementation}
The ZOFedHT algorithm is given in \Cref{alg:pseudocode}. Apart from the initial solution $x_0$, the algorithm receives several additional hyperparameters: a sequence of step-sizes $\eta_r$, a smoothing radius $\mu$, a positive integer $\tau$ denoting the length of historical trajectories, a constant $K$ denoting the number of local updates, and a combination coefficient $\alpha\in [0,1)$ used for constraining the covariance matrix.

The optimization loop of ZOFedHT follows that of FedAvg, but differs in 1) the use of non-isotropic Gaussian smoothing to estimate gradients and 2) the additional communication/computation for building the non-isotropic Gaussian covariance. Precisely, we divide the optimization loop into several rounds, and in the $r$-th round, we specify a covariance matrix $C_r$ to estimate gradients on the client side. This matrix is kept constant over local updates and across clients. To improve convergence, we build $C_r$ with solution trajectories collected on the server side, using the non-isotropic Gaussian smoothing method described in \Cref{sec:nonisotropic-gaussian-smoothing}. We detail the implementation below.

\begin{algorithm}[t]
\footnotesize
\caption{ZOFedHT}\label{alg:pseudocode}
\begin{algorithmic}[1]\setlength{\itemsep}{0pt}\setlength{\parskip}{0pt}
\State Input $x_0\in \mathbb{R}^n, \eta_r \in \mathbb{R}_+ , \mu \in \mathbb{R}_+, \tau \in \mathbb{Z}_+$
\For {$r \gets 0, 1, \dots, R-1$}
    \State Select a client set $\mathcal{W}_r$ randomly and uniformly with replacement
    \State Broadcast $x_r$ to clients in $\mathcal{W}_r$
    \If {$r \in \{\tau, 2\tau, \dots\}$}
        \State $Q_r = \text{QR}([\Delta_{r-1},\dots,\Delta_{r-\tau}])$
        \State Broadcast $Q_r$ to clients in $\mathcal{W}_r$
    \EndIf
    \For {each client $i \in \mathcal{W}_r$}
        \State $C_r \gets \begin{cases}
            I, &\text{ if } r < \tau \\ 
            (1-\alpha)I + \alpha Q_r Q_r^\top, &\text{ if } r \in \{\tau,2\tau,\dots\} \\
            C_{r-1} &\text{ otherwise } 
        \end{cases}$
        \State $w_{r,0}^i \gets x_r$
        \For {$k \gets 0, 1, ..., K-1$}
            \State $v_{r,k}^i \sim \mathcal{N}(0, C_r)$
            \State $\xi_{r,k}^i \sim \mathcal{D}_i$
            \State $g_{r,k}^i = \frac{F(w_{r,k}^i+\mu v_{r,k}^i; \xi_{r,k}^i)-F(w_{r,k}^i-\mu v_{r,k}^i; \xi_{r,k}^i)}{2\mu} v_{r,k}^i$
            \State $w_{r,k+1}^i = w_{r,k}^i - \eta_r g_{r,k}^i$
        \EndFor
        \State Upload $w_{r,K}^i$ to the server
    \EndFor
    \State $x_{r+1} = \frac{1}{|\mathcal{W}_r|}\sum_{i \in \mathcal{W}_r} w_{r,K}^i$
    \State $\Delta_r = x_{r+1} - x_r$
\EndFor

\end{algorithmic}
\end{algorithm}

At round $r$, the server samples randomly and uniformly a set of clients, denoted by $\mathcal{W}_r$, and broadcasts the solution $x_r$ to these clients (lines 3-4). 
If $r$ is a multiple of $\tau$, we collect the latest $\tau$ solution trajectories $\Delta_{r-1},\dots,\Delta_{r-\tau}$, where $\Delta_j$ denotes the change of the server-side solution computed at the end of round $j$ (line 21).
We apply QR factorization on a matrix holding these trajectories as columns (line 6), as it provides a computationally efficient and numerically highly accurate approach for this orthogonal decomposition scenario. 
This yields a $\tau$-dimensional subspace, encoded by an orthonormal basis matrix $Q_r \in \mathbb{R}^{n \times \tau}$, which captures the principal directions of the latest $\tau$ solution trajectories.

On the client side, when $r<\tau$, we set $C_r$ to $I_n$, which means that standard Gaussian smoothing is used in the first $\tau$ rounds since not enough solution trajectories are available. When the round index $r$ is a multiple of $\tau$, only the latest $Q_r$ is broadcast to the selected clients $\mathcal{W}_r$ (line 7), instead of the full covariance matrix $C_r$. This allows all local updates to use the most up-to-date projection information while reducing communication overheads. The clients then recover $C_r$ by linearly combining $Q_rQ_r^\top$ with the identity matrix $I_n$ according to \cref{eq:covariance-matrix-model} (line 10). When $r$ is not a multiple of $\tau$, the covariance matrix remains unchanged.

After building the covariance matrix $C_r$, the clients perform local updates using Gaussian smoothing based zeroth-order optimization (lines 11-17). In client $i$, we denote the solution at round $r$ and before the $k$-th local update by $w_{r,k}^i$.
At the $k$-th local update, we first draw a Gaussian vector $v_{r,k}^i$ from the non-isotropic Gaussian $\mathcal{N}(0,C_r)$ (line 13) and 
a random sample $\xi_{r,k}^i$ 
from the local data distribution $\mathcal{D}_i$ (line 14). 
By applying the gradient estimation method \cref{eq:unbiased-gradient-estimation-of-gaussian-smoothing} to the component objective $F(\cdot;\cdot)$, 
we obtain a gradient estimation $g_{r,k}^i$ (line 15). 
The solution is then updated along the negative gradient direction with a step-size $\eta_r$ fixed during round $r$.
After $K$ steps of the local updates, all clients in $\mathcal{W}_r$ update their local solutions to the server, and the server obtains a new global solution via aggregation (lines 18-20).

\subsection{Complexity} 
Since the covariance matrix $C_r$ is not built from scratch but instead recovered from the projection matrix $Q_r \in \mathbb{R}^{n\times \tau}$, only $n\tau$ additional entries need to be broadcast to the clients (line 7).
In addition, the broadcast is performed at every $\tau$ rounds, so the communication overhead is $\mathcal{O}(n)$ per-round, which is insignificant compared to the standard FedAvg or its zeroth-order implementations. 

Computation overheads yield on both the client side and the server side. On the server side, the thin QR factorization takes $\mathcal{O}(n\tau^2)$ time for every $\tau$ rounds, so the averaged time complexity is $\mathcal{O}(n\tau)$. On the client side, additional computation is caused by the Gaussian sampling (line 13). Recall that the covariance matrix takes the form of $C_r = (1-\alpha)I_n + \alpha Q_r Q_r^\top$, a sample drawn from $\mathcal{N}(0,C_r)$ can be decomposed into two parts as: 
\[
v \sim \mathcal{N}(0,C_r) \Leftrightarrow v = \sqrt{1-\alpha} v_1 +\sqrt{\alpha} Q_r v_2
\]
where $v_1 \in \mathbb{R}^n$ and $v_2 \in \mathbb{R}^\tau$ are independently sampled from $\mathcal{N}(0,I_n)$ and $\mathcal{N}(0,I_\tau)$, respectively.
This means that sampling the non-isotropic Gaussian can be achieved via a matrix-vector multiplication followed by a vector addition, taking $\mathcal{O}(n\tau)$ per local update. In practice, we recommend choosing a fixed and small $\tau$, e.g., $\tau=\mathcal{O}(1)$. In this setting, ZOFedHT enjoys the same computation and communication complexity as FedAvg.

\subsection{Convergence Properties} 
We make the following assumptions regarding problem \cref{eq:problem-definition}.
\begin{assumption}\label{assumption:objective-smoothness}
    The objective function $F(x;\xi)$ is $L$-smooth in $x\in \mathbb{R}^n$ for all $\xi$.
\end{assumption}
    
\begin{assumption}\label{assumption:local-gradient-variance-bound}
    The client-side data sampling has bounded variance, i.e., there exists some constant $\sigma_l \in \mathbb{R}_+$ such that
    \[
    \begin{aligned}
        \mathbb{E}_{\xi \sim \mathcal{D}_i } \left[\|\nabla F(x;\xi) - \mathbb{E}_{\xi \sim \mathcal{D}_i}[\nabla F(x;\xi)]\|^2\right] & \le \sigma_l^2 \\
    \end{aligned}
    \] 
    holds for all $i \in \{1,\dots,N\}$ and $x\in \mathbb{R}^n$.
\end{assumption}

\begin{assumption}\label{assumption:global-gradient-bound}
    The dissimilarity between each local gradient and the global gradient is bounded, i.e., there exists some constant $\sigma_g \in \mathbb{R}_+$ such that
    \[
       \|\mathbb{E}_{\xi \sim \mathcal{D}_i}[\nabla F(x;\xi)] - \nabla f(x)\| \le \sigma_g
    \]
    holds for all $i\in \{1,\dots,N\}$ and $x\in \mathbb{R}^n$.
\end{assumption}

\begin{assumption}\label{assumption:objective-function-lower-bound}
    The global objective is bounded from below by some constant $f_*$, i.e., $f(x) \ge f_*$ for all $x\in \mathbb{R}^n$.
\end{assumption}

\Cref{assumption:objective-smoothness,assumption:local-gradient-variance-bound,assumption:objective-function-lower-bound} are customary in analyzing zeroth-order stochastic optimization algorithms. \Cref{assumption:global-gradient-bound} measures the heterogeneity of the data distribution. For example, when $\sigma_g=0$, the data become IID and the problem degenerates to the classical distributed optimization problem.

Below we characterize the convergence property of ZOFedHT w.r.t. the standard $\ell_2$ norm.

\begin{theorem}\label{theorem:convergence}
    Let \Cref{assumption:global-gradient-bound,assumption:local-gradient-variance-bound,assumption:objective-function-lower-bound,assumption:objective-smoothness} hold with $\ell_2$ norm $\|\cdot\| = \|\cdot\|_* = \|\cdot\|_2$. 
    Choose constant step-sizes 
    \begin{equation}\label{eq:constant-step-size-setting}
    \eta_r = \eta = \sqrt{\frac{(f(x_0)-f_*)M}{\rrr{72}(n+4)(\sigma_g^2+\sigma_l^2)RKL}} 
    \end{equation}
    and constant client set-sizes $|\mathcal{W}_r| = M$. Suppose the number of rounds $R$ is sufficiently large and the number of local updates satisfies $K \le n$. Then, we have
    \[
    \begin{aligned}
    &\frac{1}{R}\sum_{r=0}^{R-1}\mathbb{E}\left[\left\|\nabla f(x_r)\right\|^2\right] \\
    \;\;\;&\leq \frac{\rrr{192}}{1-\alpha}
    \sqrt{\frac{(f(x_0)-f_*)(n+4)(\sigma_g^2+\sigma_l^2)\rrr{L}}{RKM}} \\
    &\quad + \frac{8\mu^2 L^2(n+6)^3}{(1-\alpha)^2}.
    \end{aligned}
    \]
\end{theorem}

The proof and explicit step-size conditions are given in Appendix~\ref{append:proof-of-convergence-theorem}.

For fixed $L$, $\alpha$, and noise bounds, the leading term has order $\mathcal O(\sqrt{n/(RKM)})$, subject to the stated step-size restrictions. The bound also contains a smoothing residual of order $\mu^2$; sufficiently small stationarity error therefore requires a correspondingly small smoothing radius. The dependence on $R$, $K$, and $M$ is of the same order as in FedAvg analyses~\cite{reddi2020adaptive}, while the dimension dependence is characteristic of zeroth-order estimation~\cite{fang2022communication}.
The above result does not capture the effect of non-isotropic Gaussian smoothing, as the bound can be further reduced by setting $\alpha=0$. This is expected, since the analysis relies on the $\ell_2$ norm and does not exploit second-order curvature information. We demonstrate the acceleration effect empirically in the next section, and leave a tighter analysis under additional structural assumptions for future work.
\begin{figure}[bt]
    \small
    \centering
    \includegraphics[width=0.60\linewidth]{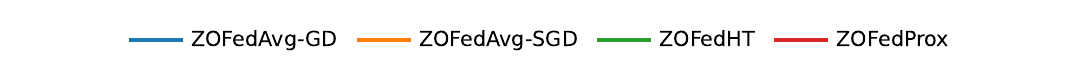}
    \vspace{-1mm}
    \subfloat[LR, mnist]{\includegraphics[trim=20 0 5 0,clip,width=0.332\linewidth]{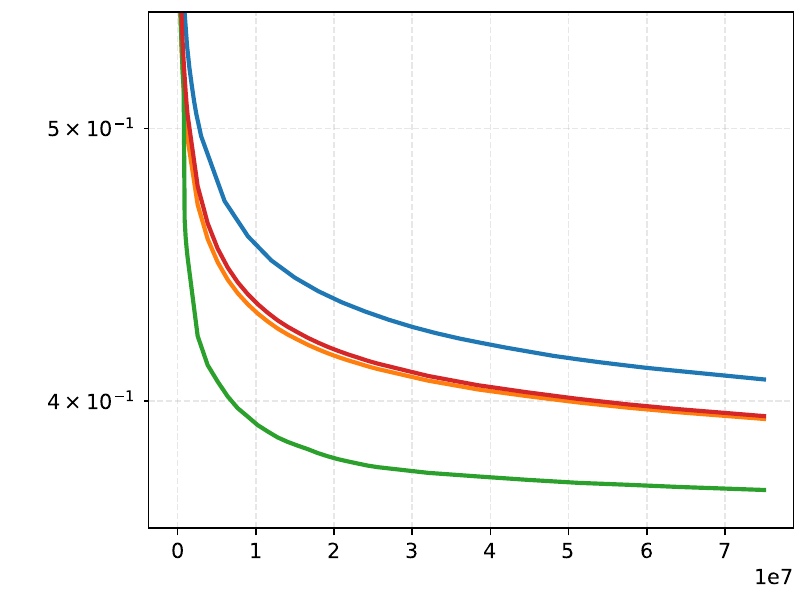}}%
    \subfloat[SVM, mnist]{\includegraphics[trim=20 0 5 0,clip,width=0.332\linewidth]{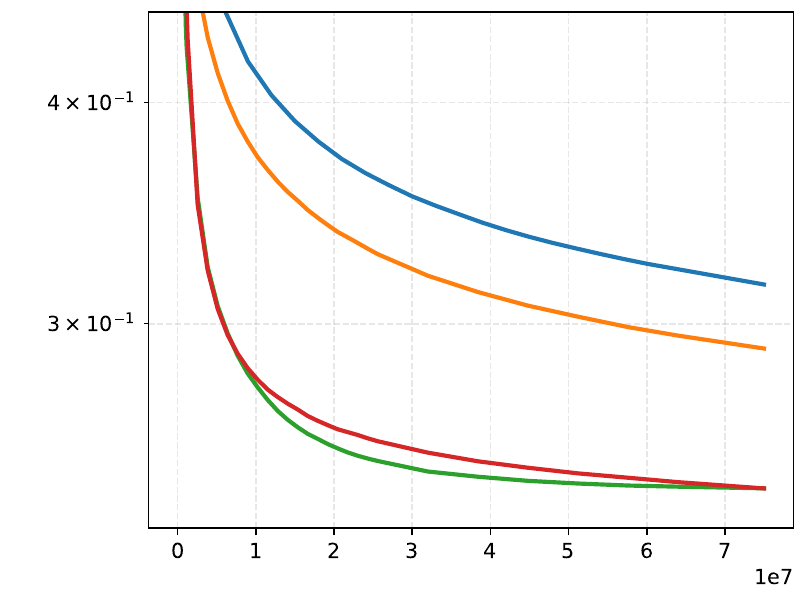}}%
    \subfloat[MLP, mnist]{\includegraphics[trim=20 0 5 0,clip,width=0.332\linewidth]{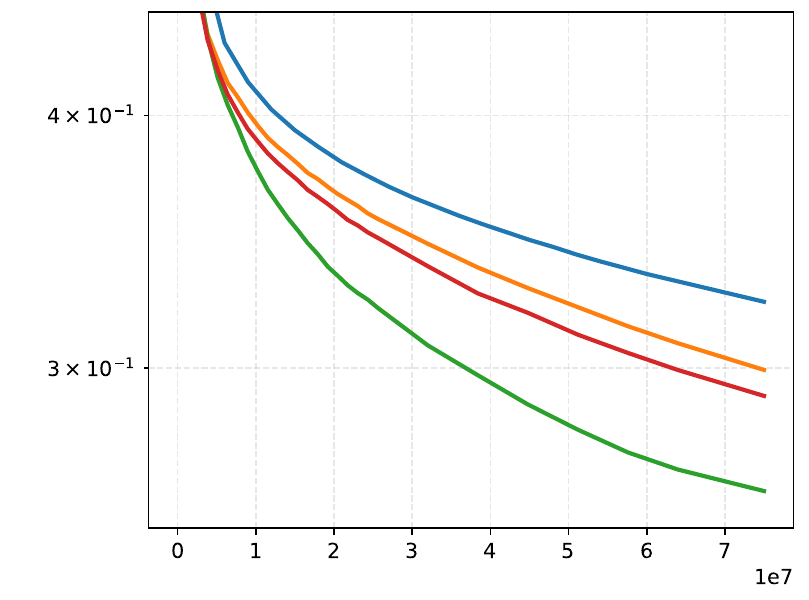}}\\[-1mm]
    \subfloat[LR, fashion-mnist]{\includegraphics[trim=20 0 5 0,clip,width=0.332\linewidth]{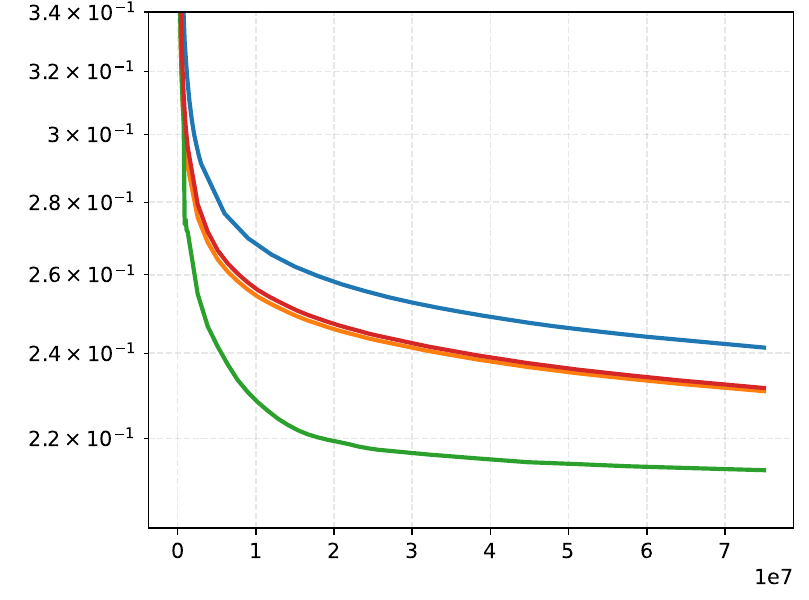}}%
    \subfloat[SVM, fashion-mnist]{\includegraphics[trim=20 0 5 0,clip,width=0.332\linewidth]{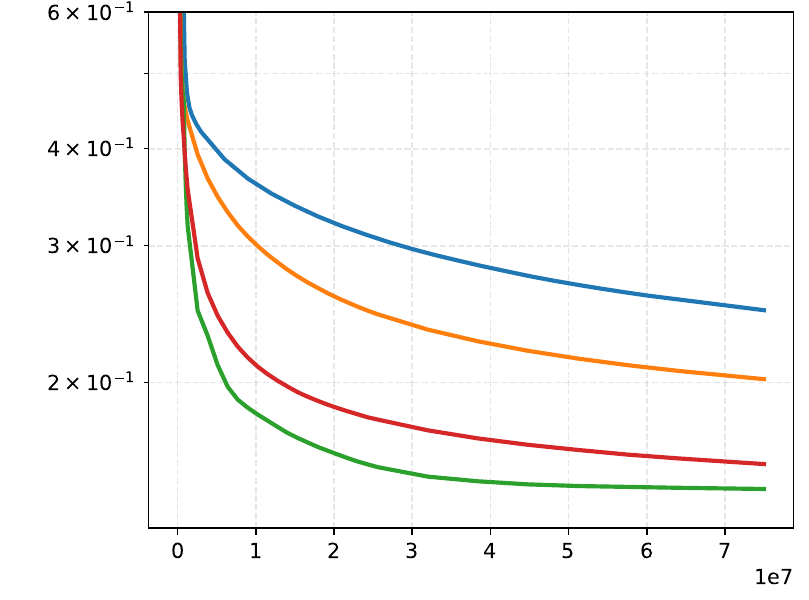}}%
    \subfloat[MLP, fashion-mnist]{\includegraphics[trim=20 0 5 0,clip,width=0.332\linewidth]{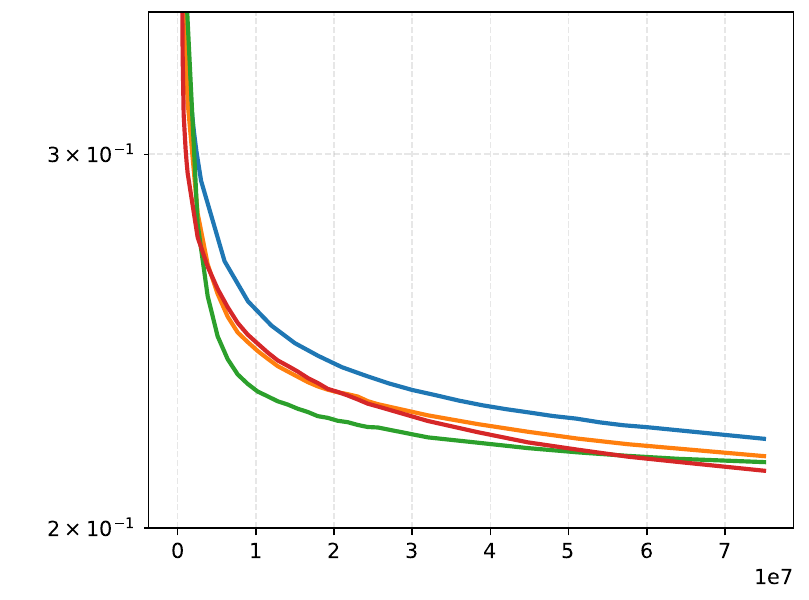}}\\[-1mm]
    \subfloat[LR, rcv1]{\includegraphics[trim=20 0 5 0,clip,width=0.332\linewidth]{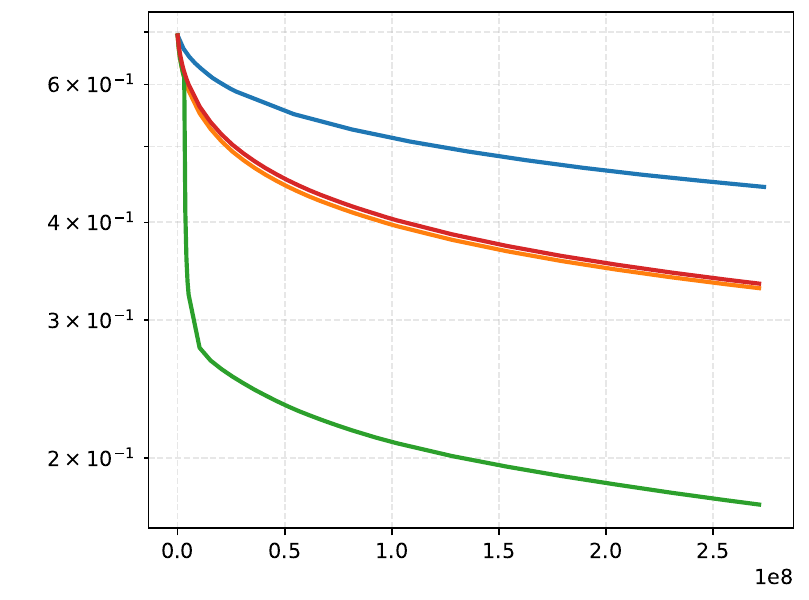}}%
    \subfloat[SVM, rcv1]{\includegraphics[trim=20 0 5 0,clip,width=0.332\linewidth]{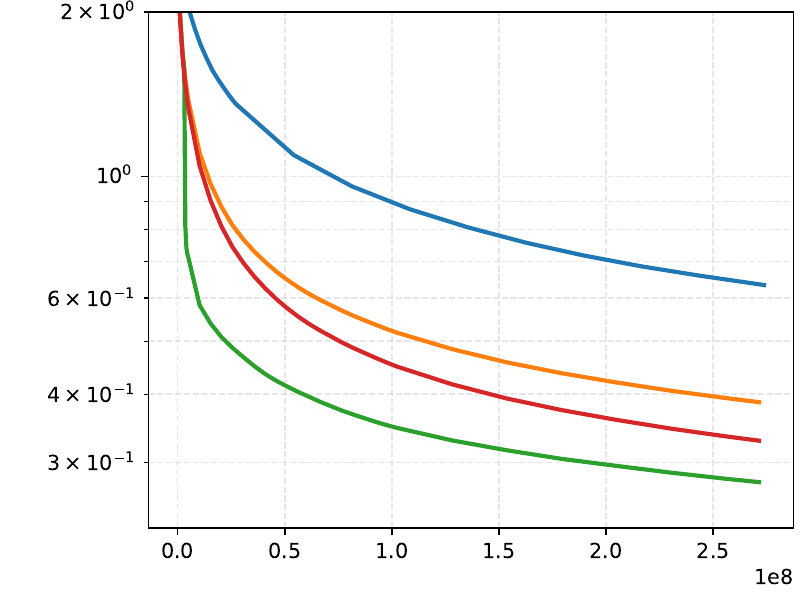}}%
    \subfloat[MLP, rcv1]{\includegraphics[trim=20 0 5 0,clip,width=0.332\linewidth]{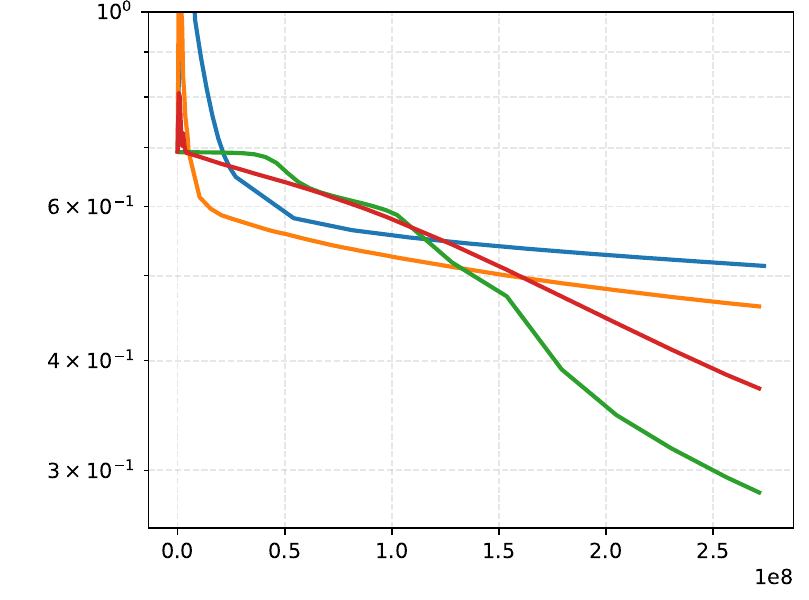}}
    \caption{Training loss versus the number of function evaluations obtained in the IID setting}
    \label{tab:training-loss-iid}
\end{figure}
\section{Numerical Studies}
\label{Experiments}
\subsection{Experimental Setup}
We evaluate ZOFedHT using logistic regression (LR), support vector machines (SVM), and multilayer perceptrons (MLP), representing smooth convex, nonsmooth convex, and smooth nonconvex objectives, respectively. The MLP has one fully connected hidden layer with 50 neurons and sigmoid activations.

We compare ZOFedHT with ZOFedAvg-SGD, ZOFedAvg-GD, and ZOFedProx. The first two are zeroth-order variants of FedAvg using stochastic and full-batch gradient estimates, respectively, whereas ZOFedProx applies standard Gaussian smoothing to FedProx~\cite{li2020federated}.

In all algorithms, the smoothing radius is $\mu=10^{-4}$ and the number of local updates is $K=50$. 
All algorithms use step-sizes decreasing over rounds as $\eta_r = \eta_0/\sqrt{r+1}$, where the initial learning rate $\eta_0$ is tuned via grid search over $\{0.1, 1.0, 10\}$ for each algorithm and dataset, and the best-performing one is selected.

Three widely used benchmark datasets, including mnist~\cite{deng_mnist_2012}, fashion-mnist~\cite{xiao2017online}, and rcv1~\cite{david_d_lewis_rcv1_2004}, are chosen. 
Our experiments simulate a binary classification problem using these datasets. For mnist and fashion-mnist, which contain images of digits from 0 to 9, we re-classify the samples into two categories: one with digits 0 to 4 and the other one with 5 to 9. 
We set the number of clients $N$ to 100, and at each round 10 clients are sampled uniformly. 
To account for the varying scales of these datasets, we employ a dataset-specific batch size strategy for the client-side data sampling in ZOFedHT, ZOFedProx, and ZOFedAvg-SGD. Specifically, the batch size $B$ is set to 256 for the mnist and fashion-mnist datasets, and 1024 for the significantly larger rcv1 dataset.
\begin{figure}[bt]
    \small
    \centering
    \includegraphics[width=0.60\linewidth]{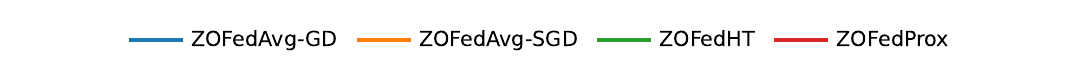}
    \vspace{-1mm}
    \subfloat[LR, mnist]{\includegraphics[trim=20 0 5 0,clip,width=0.332\linewidth]{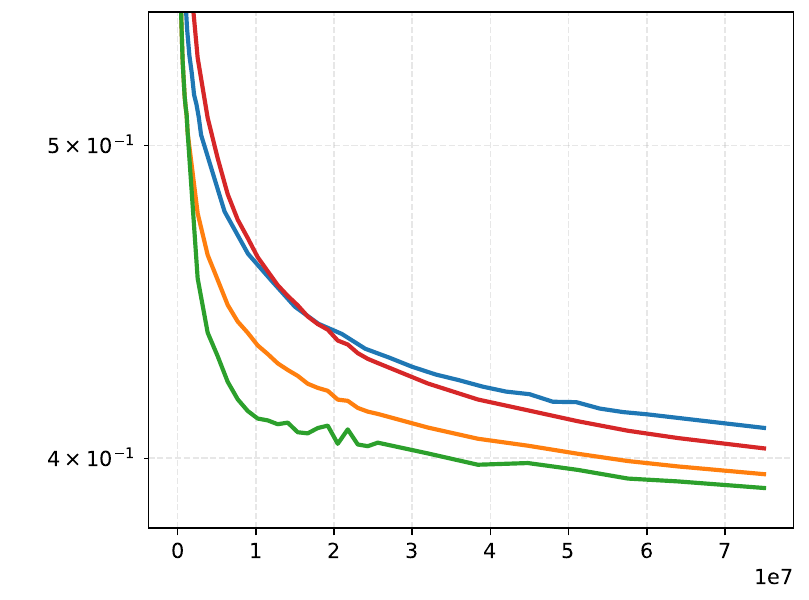}}%
    \subfloat[SVM, mnist]{\includegraphics[trim=20 0 5 0,clip,width=0.332\linewidth]{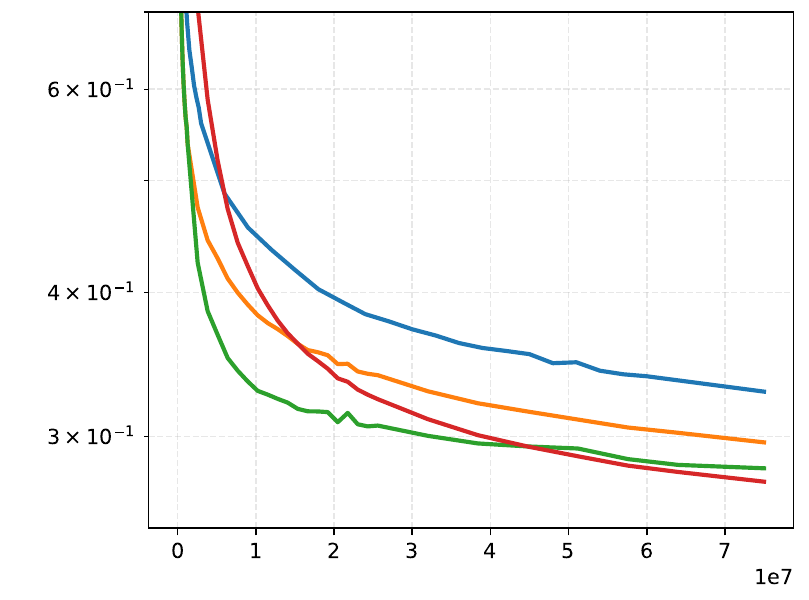}}%
    \subfloat[MLP, mnist]{\includegraphics[trim=20 0 5 0,clip,width=0.332\linewidth]{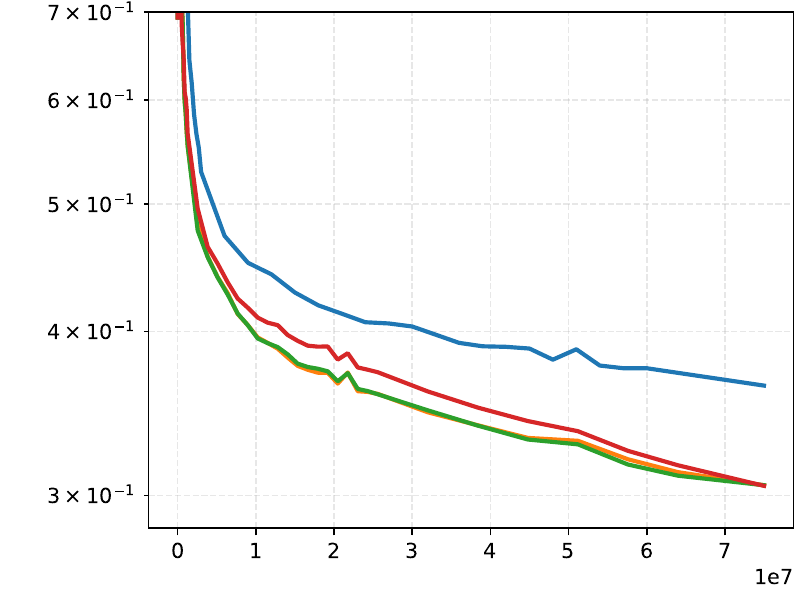}}\\[-1mm]
    \subfloat[LR, fashion-mnist]{\includegraphics[trim=20 0 5 0,clip,width=0.332\linewidth]{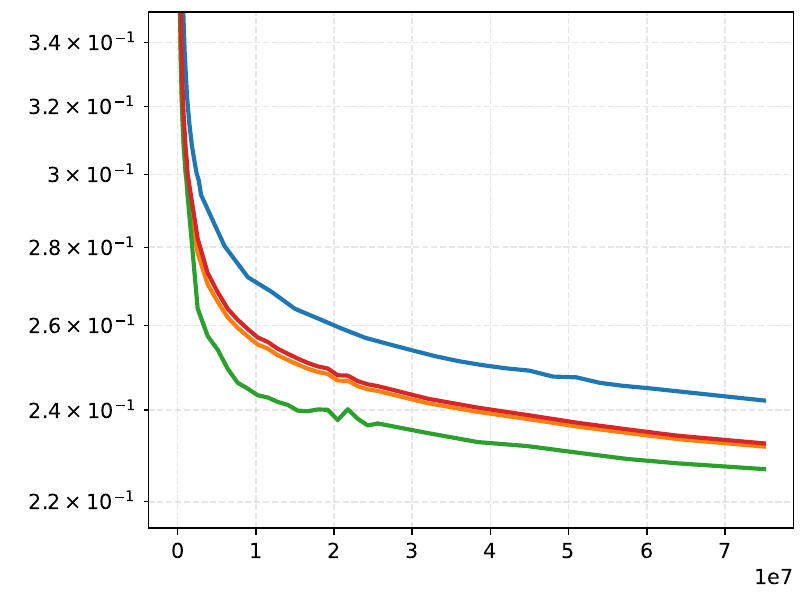}}%
    \subfloat[SVM, fashion-mnist]{\includegraphics[trim=20 0 5 0,clip,width=0.332\linewidth]{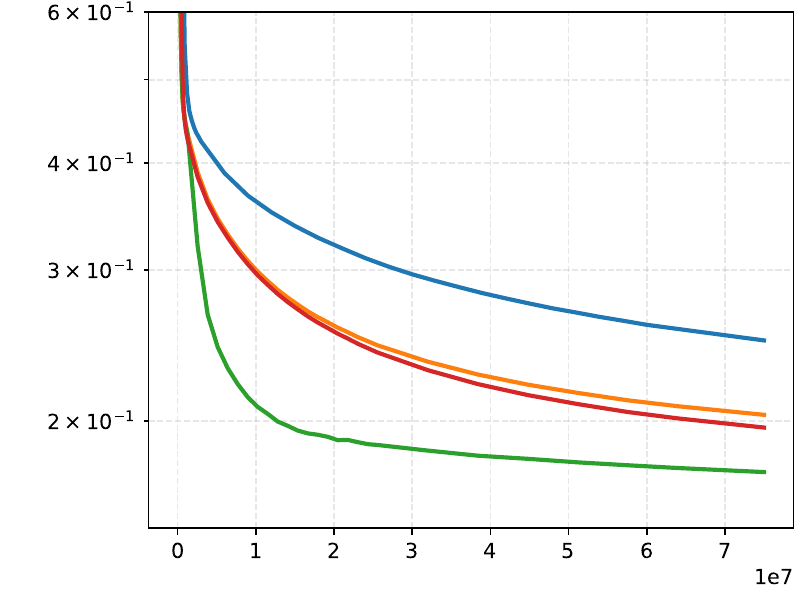}}%
    \subfloat[MLP, fashion-mnist]{\includegraphics[trim=20 0 5 0,clip,width=0.332\linewidth]{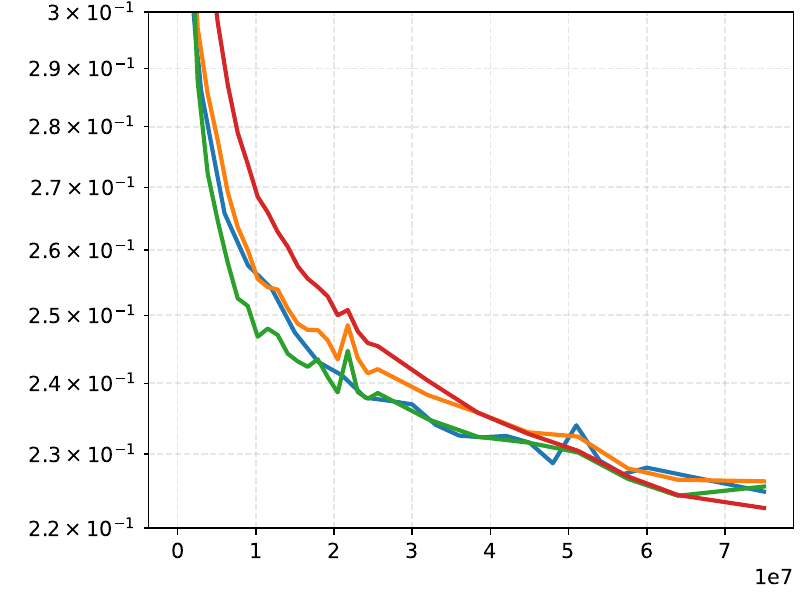}}\\[-1mm]
    \subfloat[LR, rcv1]{\includegraphics[trim=20 0 5 0,clip,width=0.332\linewidth]{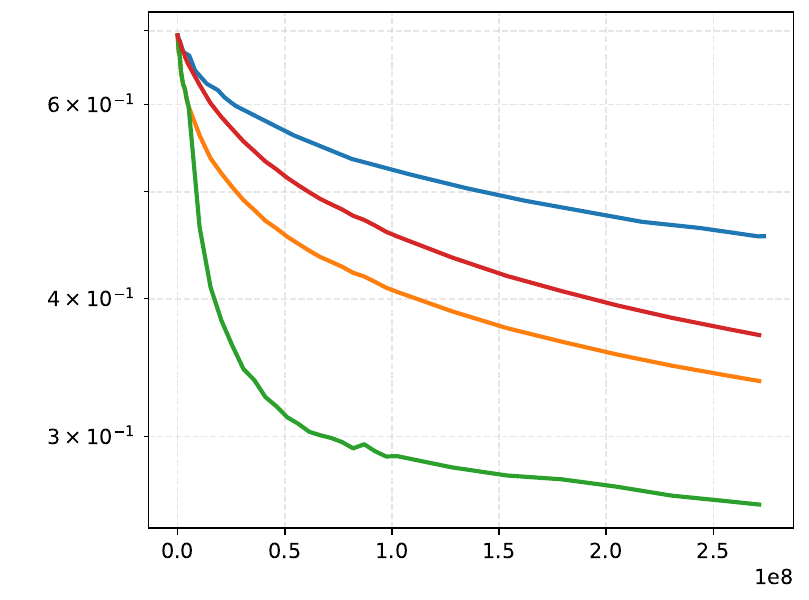}}%
    \subfloat[SVM, rcv1]{\includegraphics[trim=20 0 5 0,clip,width=0.332\linewidth]{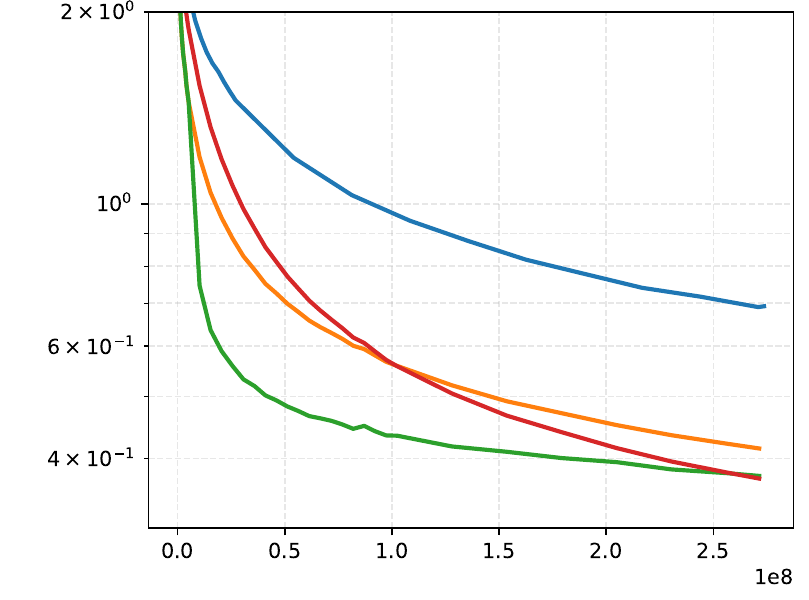}}%
    \subfloat[MLP, rcv1]{\includegraphics[trim=20 0 5 0,clip,width=0.332\linewidth]{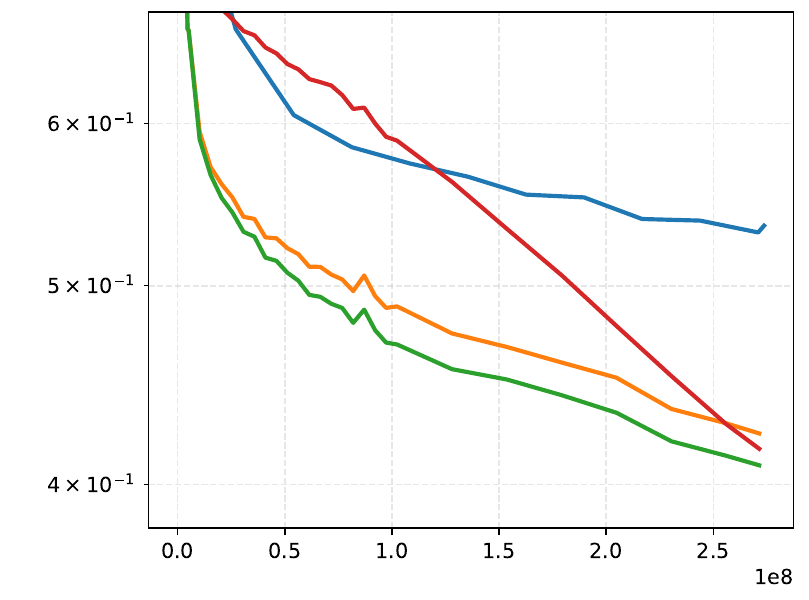}}
    \caption{Training loss versus the number of function evaluations obtained in the non-IID setting}
    \label{tab:training-loss-noniid}
\end{figure}

\subsection{Main Results }
For all IID tasks, we fix $\tau=5$ and $\alpha=0.3$, while for all non-IID tasks, we fix $\tau=10$ and $\alpha=0.1$. To ensure statistical rigor, all methods are run three times with different random seeds, and the reported results are averaged over these runs.

For IID experiments, the data are randomly and uniformly partitioned among $N$ clients.
\Cref{tab:training-loss-iid} shows that ZOFedHT achieves the best or competitive performance across the evaluated tasks, demonstrating the effectiveness of non-isotropic Gaussian smoothing.

For non-IID experiments, the samples are sorted by label and evenly partitioned, so each client may observe only a subset of labels.
\Cref{tab:training-loss-noniid} shows that ZOFedHT generally converges faster and obtains lower or comparable final training losses under data heterogeneity.

Together with the IID results in \Cref{tab:training-loss-iid}, these observations demonstrate that the trajectory-assisted sampling strategy remains effective under both homogeneous and heterogeneous client-data distributions.
\subsection{Sensitivity Analysis of \texorpdfstring{$\tau$}{tau} and \texorpdfstring{$\alpha$}{alpha}}
\begin{figure}[htbp]
    \small
    \centering
    \subfloat[IID]{\includegraphics[width=0.48\linewidth]{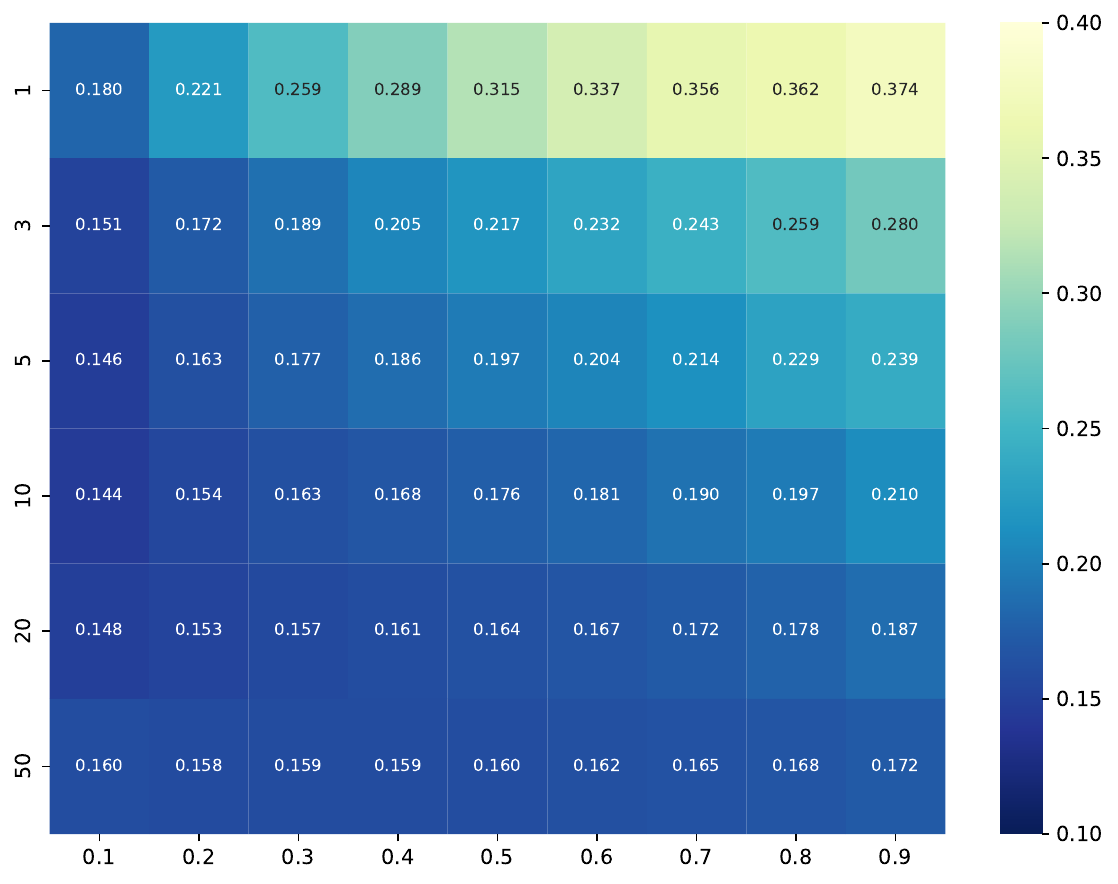}}
    \hfill
    \subfloat[Non-IID]{\includegraphics[width=0.48\linewidth]{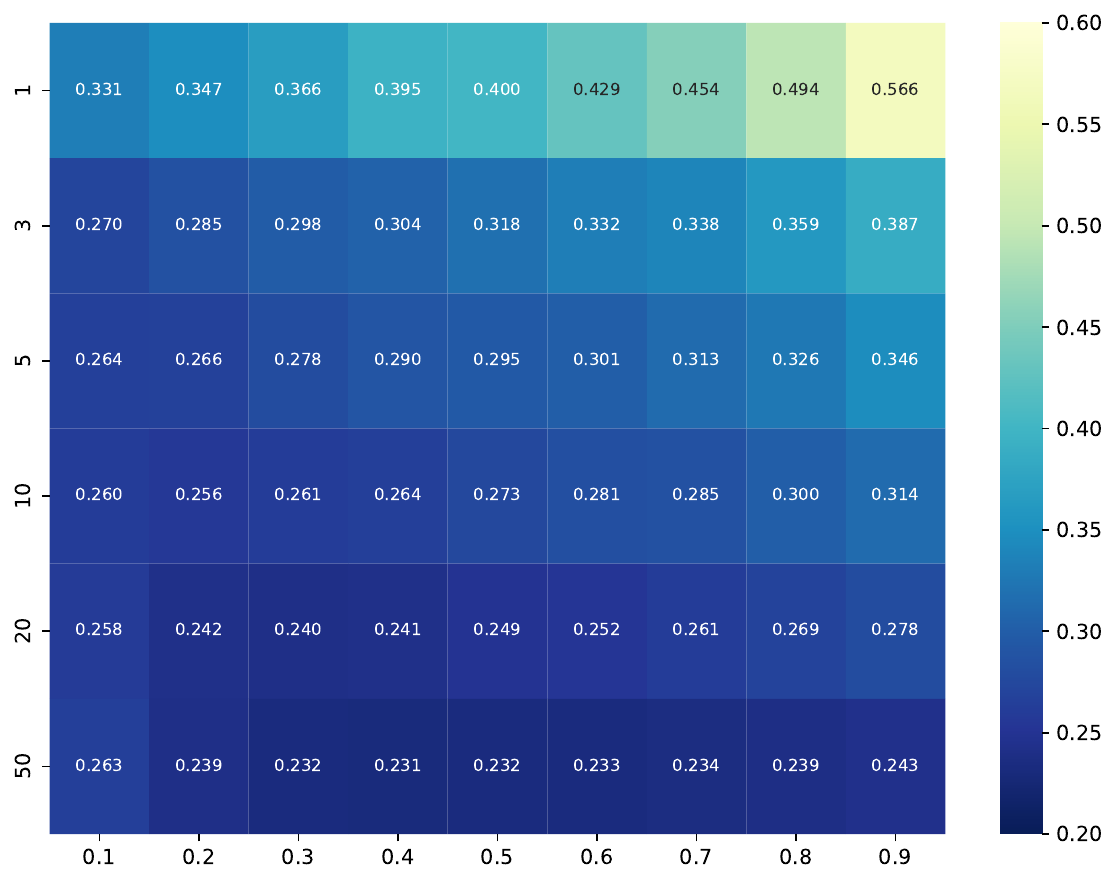}}
    \caption{Sensitivity analysis of ZOFedHT on logistic regression over the rcv1 dataset. The horizontal axis denotes $\alpha$ ranging from 0.1 to 0.9, the vertical axis denotes $\tau \in \{1,3,5,10,20,50\}$, and each entry reports the final test loss}
    \label{fig:sensitivity_analysis}
\end{figure}

ZOFedHT introduces two additional hyperparameters in the non-isotropic Gaussian smoothing, namely $\alpha$ and $\tau$.
We conduct a sensitivity analysis on a representative task, namely logistic regression on the rcv1 dataset. As shown in \Cref{fig:sensitivity_analysis}, the performance remains relatively stable over a broad range of $\alpha$ and $\tau$. In particular, as long as $\tau$ is not too small and $\alpha$ is not too large, the final test loss stays at a similar level in both IID and non-IID settings. This suggests that ZOFedHT is insensitive to the exact choice of $\alpha$ and $\tau$.

\label{sec:sensitivity-analysis}

\section{Related Work}
\label{sec:related-work}
Gaussian smoothing is perhaps the most well-known technique for gradient estimation~\cite{ghadimi2013stochastic,nesterov2017random}. Although using an identity covariance matrix could be effective in certain cases~\cite{balasubramanian_zeroth-order_2022-1}, it has been long recognized that incorporating second-order information in Gaussian smoothing could enhance convergence.
For example, \cite{he2022adaptive} suggested that, for convex quadratic problems, the optimal covariance matrix in Gaussian smoothing should be proportional to the inverse of the Hessian, coinciding with our statement in \Cref{ss:covariance-matrix-specification}.
Several approaches exist for exploiting Hessian information when using Gaussian smoothing.
\cite{bollapragada_adaptive_2023} proposed a quasi-Newton procedure for extracting the curvature information from gradient estimators output by Gaussian smoothing.
\cite{zhang_zeroth-order_2022} proposed a four-point rule for estimating the Hessian-vector product, which mimics the two-point rule in Gaussian smoothing.
\cite{ye_hessian-aware_2019} suggested estimating the Hessian explicitly and using the inverse of the estimated Hessian as the covariance matrix.

While improving Gaussian smoothing in convex problems, using Hessian information might cause divergence in non-convex cases. 
A workaround is to exploit historical trajectories instead of the Hessian information.
One notable example is the guided evolution strategy method for reinforcement learning~\cite{maheswaranathan2019guided}.
It estimates gradients in the parameter space using Gaussian smoothing, and explores the principal components of latest gradients to bias the sampling procedure. 
\cite{liu2020self} further improves this method via self-adapting the importance of subspace sampling. 
In this paper we extend these methods to federated settings and provide theoretical guarantees for those methods using historical trajectories.

Zeroth-order federated methods include FedZO~\cite{fang2022communication}, the full-batch Newton-type FedZen~\cite{maritan2023fedzen}, and a bilevel extension of Gaussian smoothing~\cite{qiu2023zeroth}. Shu et al.~\cite{shu2023federated} also exploit historical trajectories, but use them for variance reduction rather than for adapting the smoothing covariance.

\section{Conclusion}
\label{conclusion}
We proposed ZOFedHT, which uses historical global trajectories to construct a non-isotropic smoothing covariance for zeroth-order federated optimization. Under the stated sampling and step-size conditions, it retains the same-order leading convergence rate as existing zeroth-order federated methods, with an explicit smoothing residual. It has the same-order computation and communication complexity as the baselines and performs competitively in the experiments. For extremely large models, the additional communication of $Q_r$ could be reduced through quantization.

\appendix
\raggedbottom
\crefalias{section}{appendix}
\section{Additional Quantitative Analysis and Test Performance}
\label{appendix:test_performance}
To address the quantitative evaluation of the proposed ZOFedHT algorithm and further assess its generalization performance, we present the test performance corresponding to the training trajectories discussed in~\Cref{Experiments}. 

\Cref{tab:test_loss_comparison} provides a quantitative comparison of the final test losses achieved by different algorithms under both IID and non-IID settings. 
The results clearly indicate that ZOFedHT 
achieves the lowest test loss in 14 of the 18 scenarios and remains competitive in the others.
\begin{figure}[htbp]
    \small
    \centering
    \includegraphics[width=0.7\linewidth]{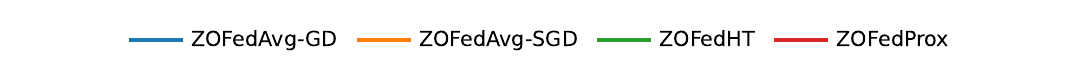}
    \vspace{0mm}
    \subfloat[LR, mnist]{\includegraphics[width=0.33\linewidth]{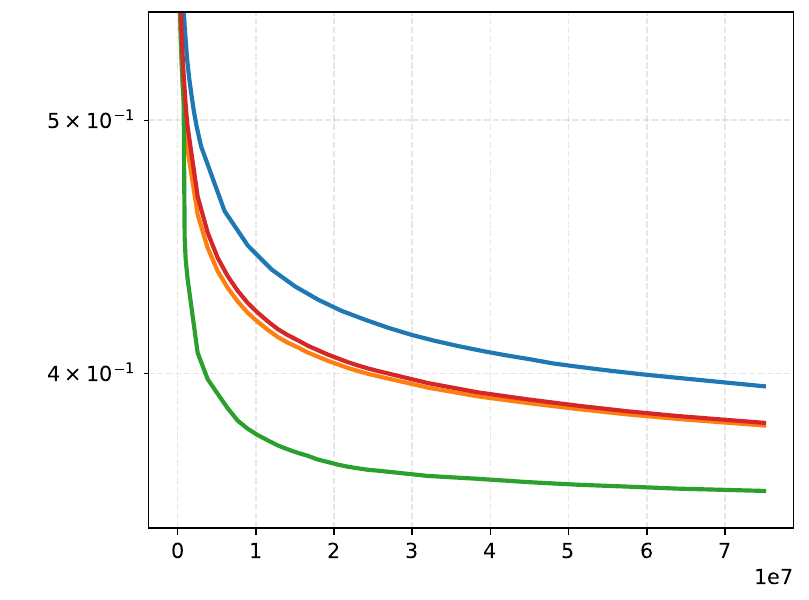}}
    \subfloat[SVM, mnist]{\includegraphics[width=0.33\linewidth]{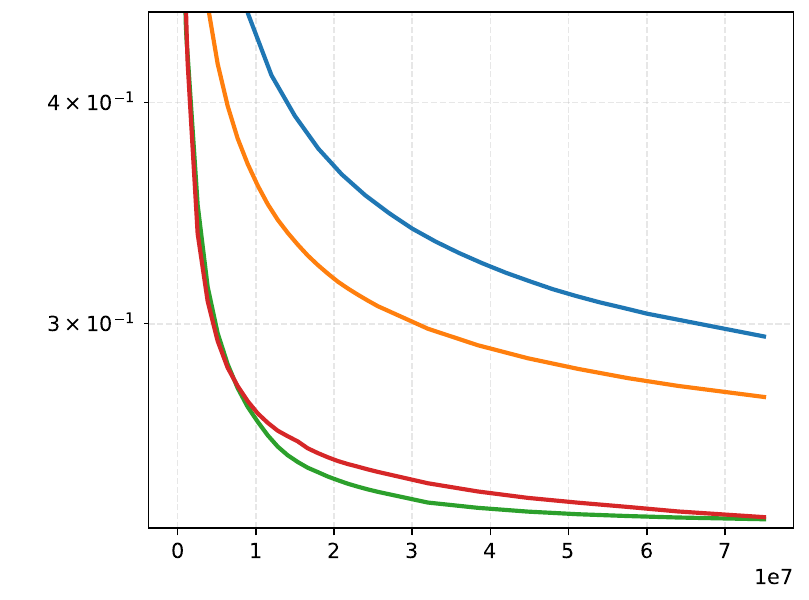}}
    \subfloat[MLP, mnist]{\includegraphics[width=0.33\linewidth]{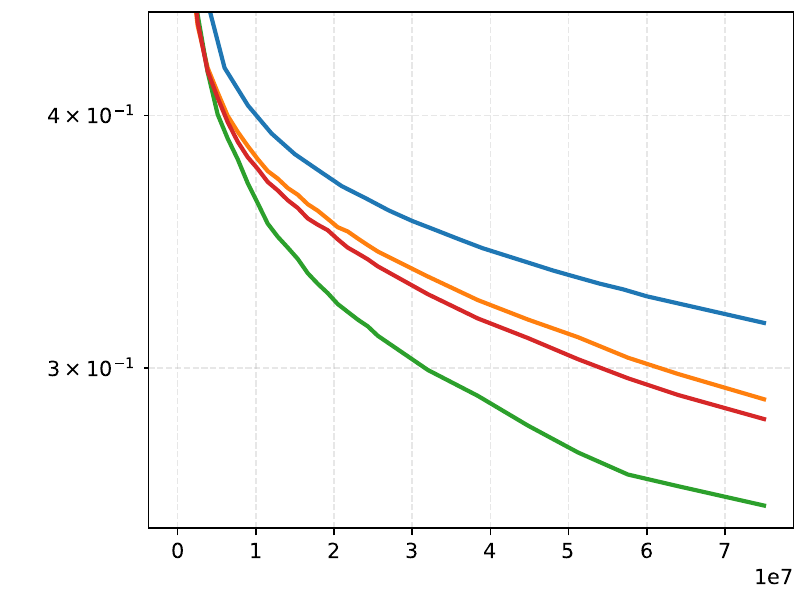}}\\
    \subfloat[LR, fashion-mnist]{\includegraphics[width=0.33\linewidth]{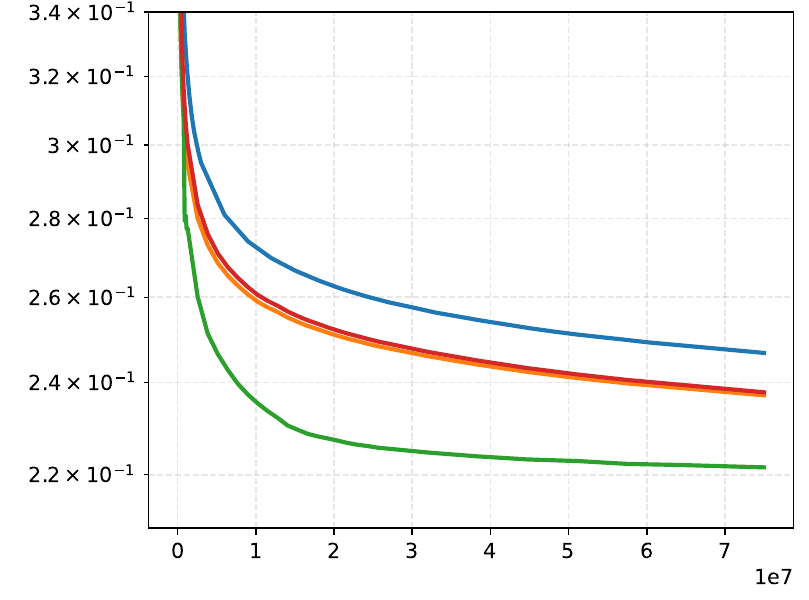}}
    \subfloat[SVM, fashion-mnist]{\includegraphics[width=0.33\linewidth]{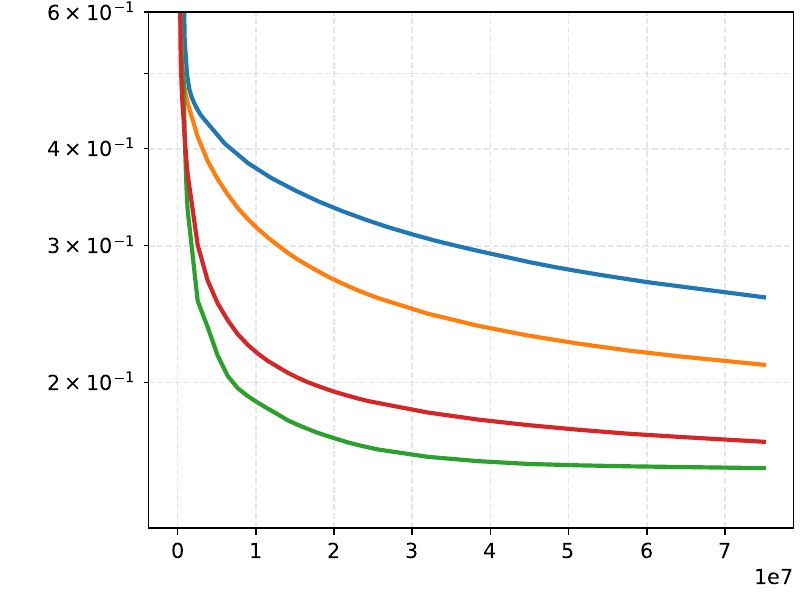}} 
    \subfloat[MLP, fashion-mnist]{\includegraphics[width=0.33\linewidth]{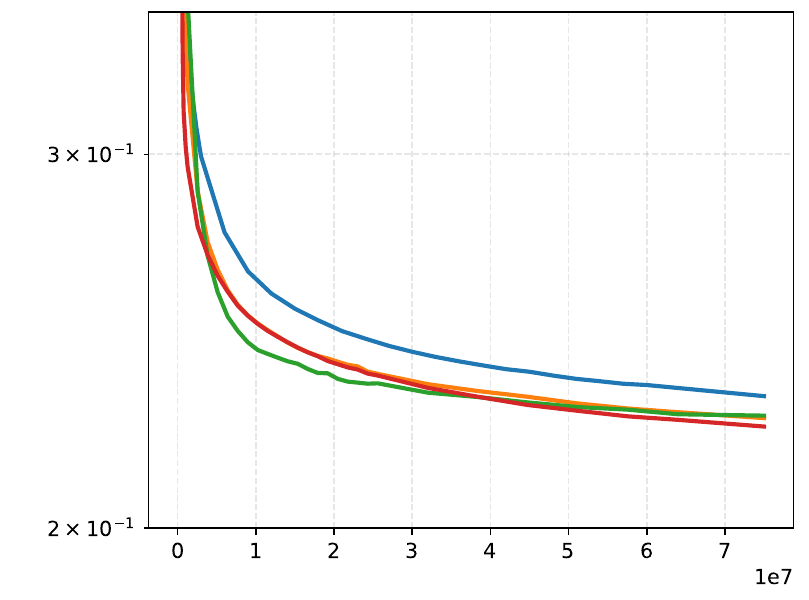}}\\
    \subfloat[LR, rcv1]{\includegraphics[width=0.33\linewidth]{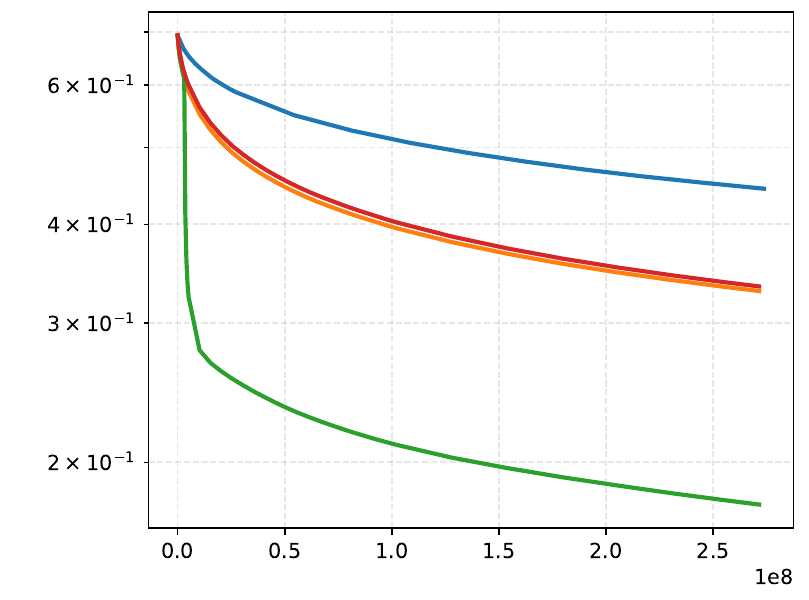}}
    \subfloat[SVM, rcv1]{\includegraphics[width=0.33\linewidth]{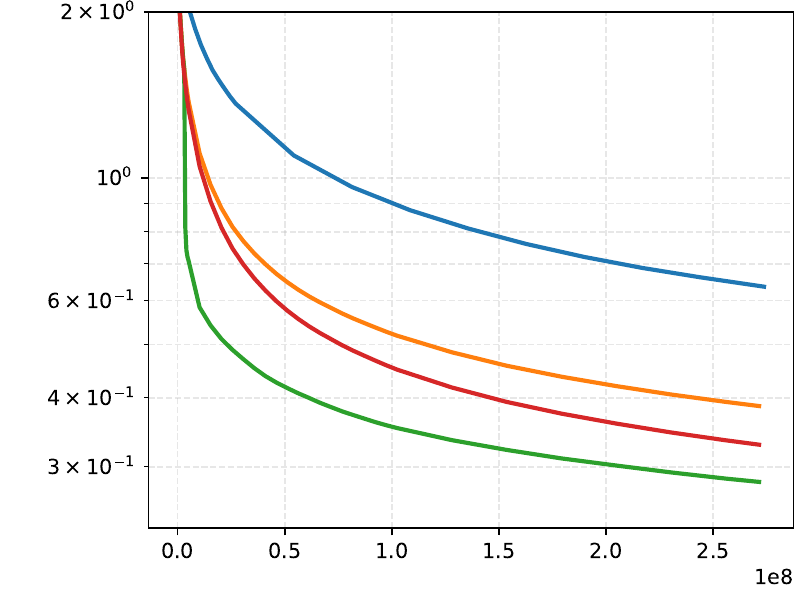}}
    \subfloat[MLP, rcv1]{\includegraphics[width=0.33\linewidth]{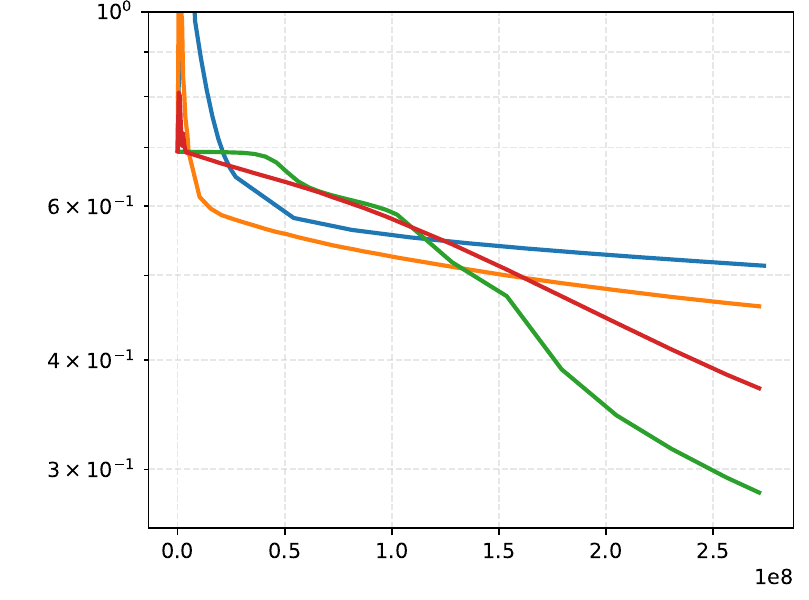}}
    \caption{Test loss versus the number of function evaluations obtained in the IID setting}
    \label{fig:test-loss-iid}
\end{figure}
\begin{figure}[htbp]
    \small
    \centering
    \includegraphics[width=0.7\linewidth]{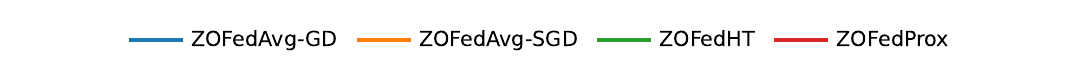}
    \vspace{0mm}
    \subfloat[LR, mnist]{\includegraphics[width=0.33\linewidth]{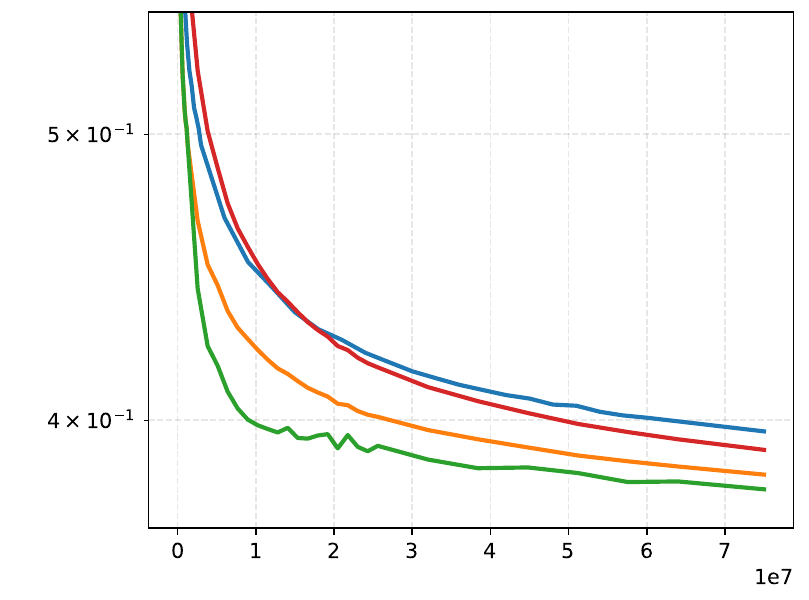}}
    \subfloat[SVM, mnist]{\includegraphics[width=0.33\linewidth]{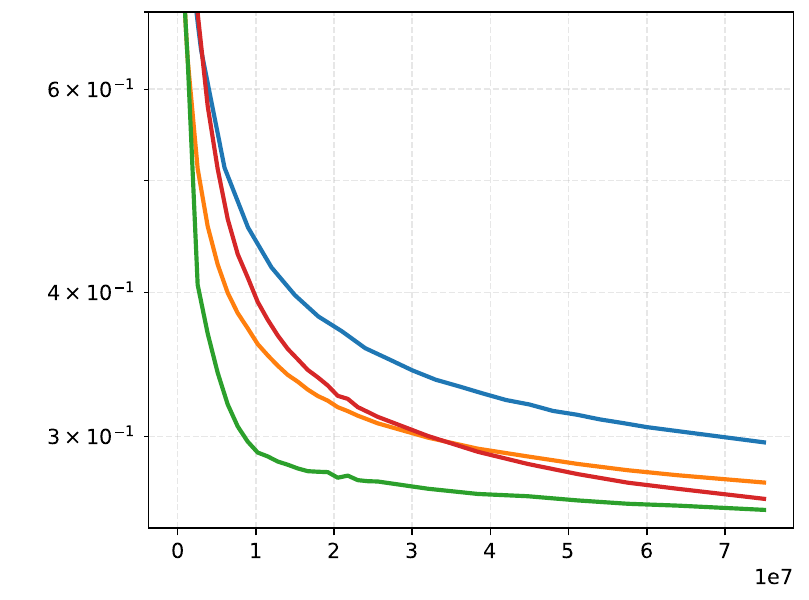}}
    \subfloat[MLP, mnist]{\includegraphics[width=0.33\linewidth]{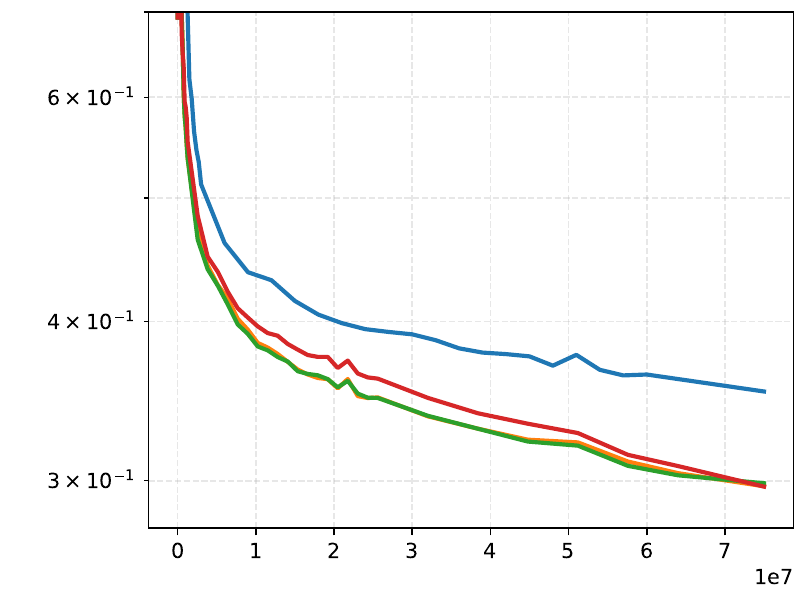}}\\
    \subfloat[LR, fashion-mnist]{\includegraphics[width=0.33\linewidth]{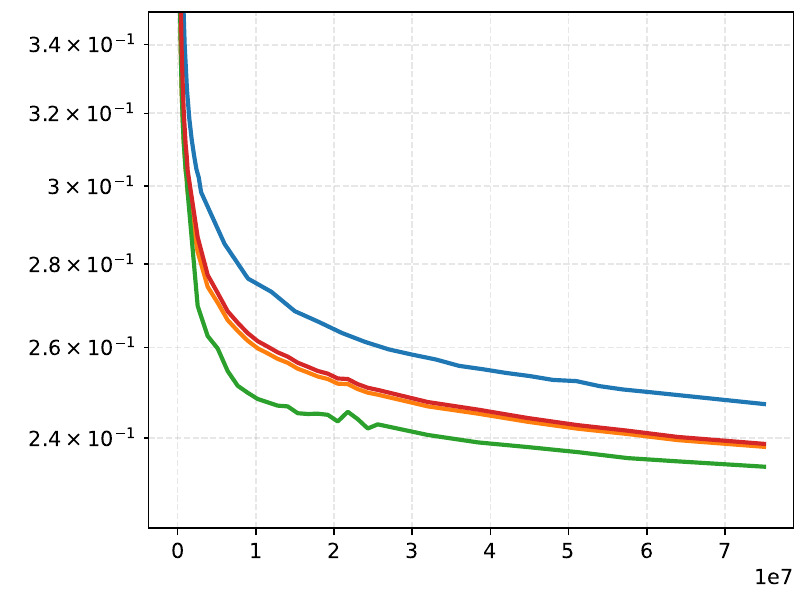}}
    \subfloat[SVM, fashion-mnist]{\includegraphics[width=0.33\linewidth]{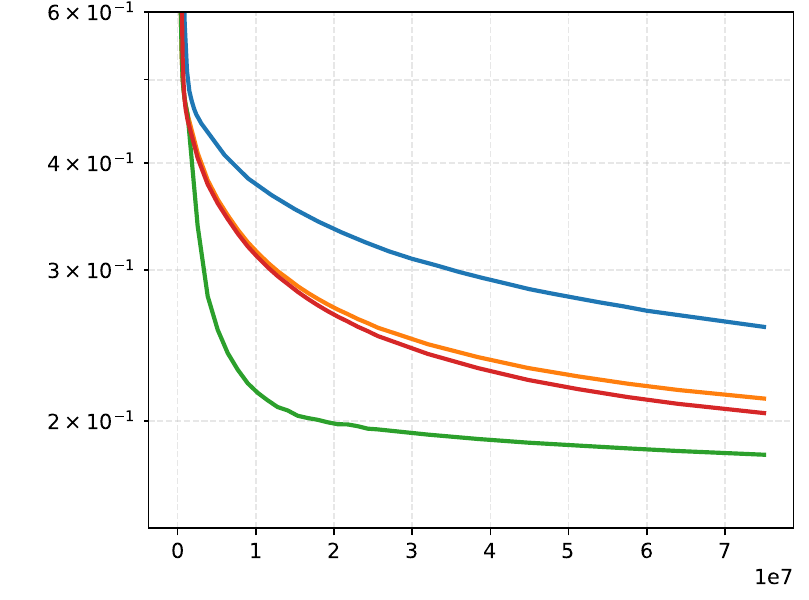}} 
    \subfloat[MLP, fashion-mnist]{\includegraphics[width=0.33\linewidth]{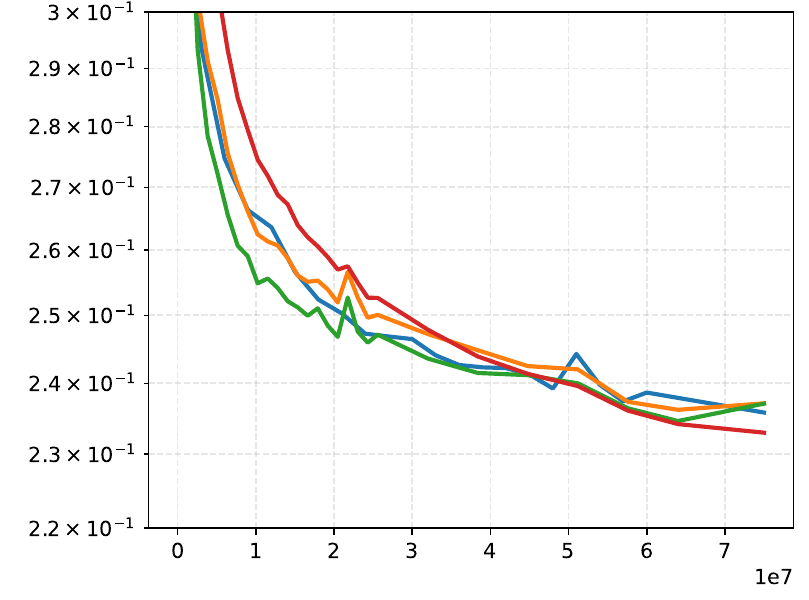}}\\
    \subfloat[LR, rcv1]{\includegraphics[width=0.33\linewidth]{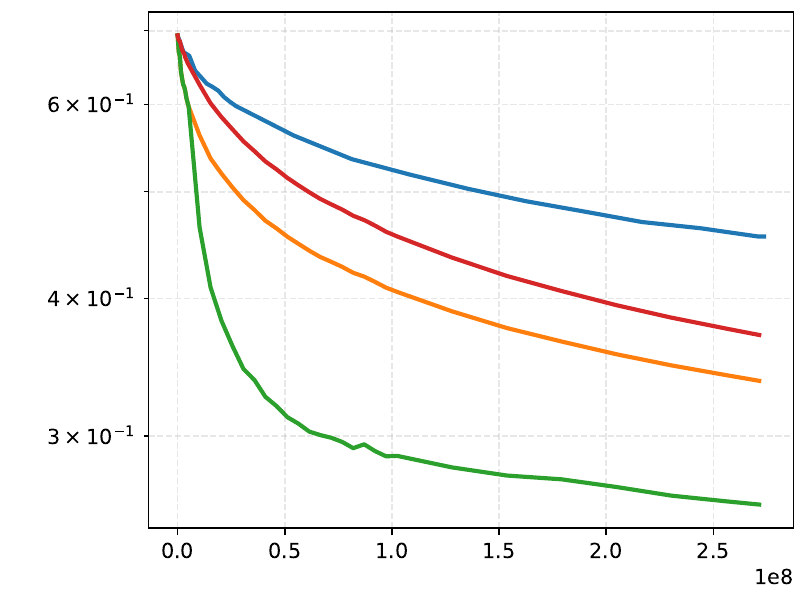}}
    \subfloat[SVM, rcv1]{\includegraphics[width=0.33\linewidth]{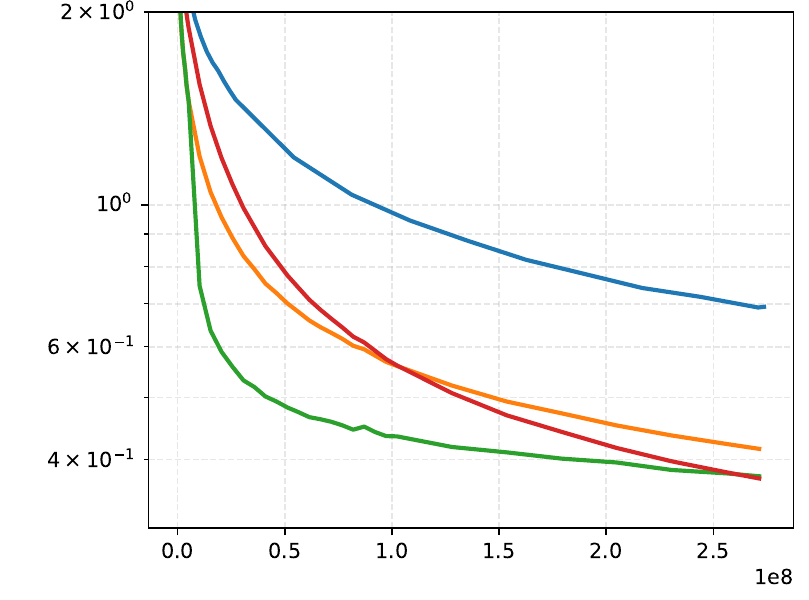}}
    \subfloat[MLP, rcv1]{\includegraphics[width=0.33\linewidth]{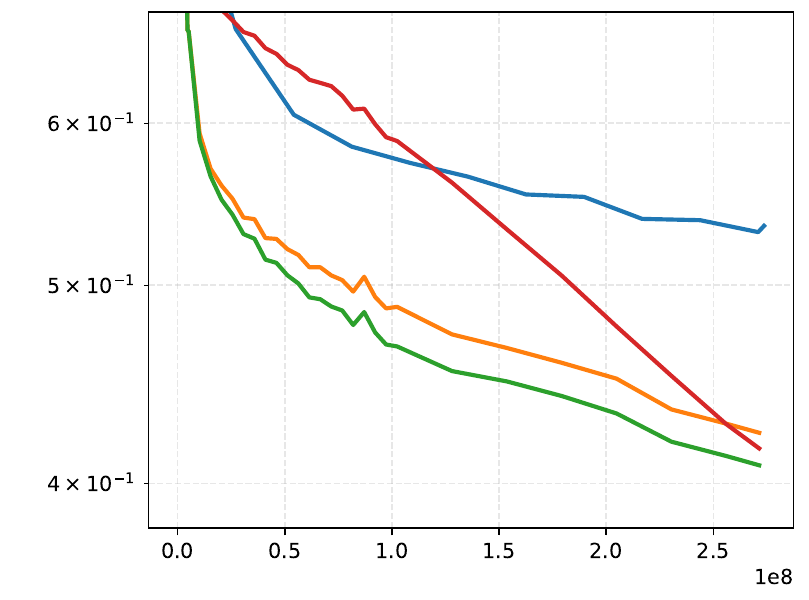}}
    \caption{Test loss versus the number of function evaluations obtained in the non-IID setting}
    \label{fig:test-loss-non-iid}
\end{figure}

Furthermore, the complete test loss trajectories over the number of function evaluations are plotted in~\Cref{fig:test-loss-iid} for the IID setting and in~\Cref{fig:test-loss-non-iid} for the non-IID setting. 
The test curves generally follow the trends observed in the training curves, suggesting that the accelerated convergence of ZOFedHT translates effectively to test performance across these evaluated tasks.

\begin{table}[htbp]
\centering
\scriptsize
\setlength{\tabcolsep}{3pt}
\caption{Comparison of final test loss under IID and non-IID settings. Lower values indicate better performance}
\label{tab:test_loss_comparison}
\resizebox{\textwidth}{!}{%
\begin{tabular}{@{}lcccccccc@{}}
\toprule
\multirow{2}{*}{\textbf{Problem}} & \multicolumn{4}{c}{\textbf{IID}} & \multicolumn{4}{c}{\textbf{Non-IID}} \\
\cmidrule(lr){2-5} \cmidrule(lr){6-9}
& \textbf{ZOFedAvg-GD} & \textbf{ZOFedAvg-SGD} & \textbf{ZOFedHT} & \textbf{ZOFedProx} & \textbf{ZOFedAvg-GD} & \textbf{ZOFedAvg-SGD} & \textbf{ZOFedHT} & \textbf{ZOFedProx} \\
\midrule
Fashion-MNIST (SVM) & $0.2576$ & $0.2109$ & $\mathbf{0.1552}$ & $0.1678$ & $0.2575$ & $0.2124$ & $\mathbf{0.1826}$ & $0.2043$ \\
Fashion-MNIST (MLP) & $0.2307$ & $0.2253$ & $0.2259$ & $\mathbf{0.2233}$ & $0.2358$ & $0.2345$ & $0.2340$ & $\mathbf{0.2324}$ \\
Fashion-MNIST (LR)  & $0.2467$ & $0.2371$ & $\mathbf{0.2216}$ & $0.2378$ & $0.2473$ & $0.2381$ & $\mathbf{0.2340}$ & $0.2388$ \\
MNIST (SVM)         & $0.2951$ & $0.2727$ & $\mathbf{0.2326}$ & $0.2333$ & $0.2966$ & $0.2737$ & $\mathbf{0.2592}$ & $0.2650$ \\
MNIST (MLP)         & $0.3158$ & $0.2895$ & $\mathbf{0.2565}$ & $0.2830$ & $0.3525$ & $0.2970$ & $0.2986$ & $\mathbf{0.2968}$ \\
MNIST (LR)          & $0.3956$ & $0.3821$ & $\mathbf{0.3607}$ & $0.3830$ & $0.3966$ & $0.3834$ & $\mathbf{0.3790}$ & $0.3909$ \\
RCV1 (SVM)          & $0.6360$ & $0.3865$ & $\mathbf{0.2817}$ & $0.3290$ & $0.6893$ & $0.4155$ & $0.3765$ & $\mathbf{0.3735}$ \\
RCV1 (MLP)          & $0.5127$ & $0.4607$ & $\mathbf{0.2821}$ & $0.3713$ & $0.5290$ & $0.4234$ & $\mathbf{0.4081}$ & $0.4161$ \\
RCV1 (LR)           & $0.4436$ & $0.3293$ & $\mathbf{0.1768}$ & $0.3337$ & $0.4549$ & $0.3367$ & $\mathbf{0.2600}$ & $0.3706$ \\
\bottomrule
\end{tabular}
\vphantom{\begin{tabular}{l}A\end{tabular}}%
}
\end{table}

\section{Proof of Lemma 1}
By \cite[lemma 1]{he2022adaptive} we have
\[
\begin{split}
    \nabla h_{\mu,C}(\rrr{x}) = \mathbb{E}_{v\sim \mathcal{N}(0,C)}\left[\frac{h(x+\mu v)-h(x)}{\mu}C^{-1}v\right].
\end{split}
\]
Substituting $v$ with $-v$ and using the fact $v \sim \mathcal{N}(0,C) \Leftrightarrow -v \sim \mathcal{N}(0,C) $, we have
\[
\nabla h_{\mu,C}(\rrr{x}) = \mathbb{E}_{v\sim \mathcal{N}(0,C)}\left[\frac{h(x)-h(x -\mu v)}{\mu}C^{-1}v\right].
\]
\Cref{eq:gradient_gaussian_smoothing} can be obtained by summing the above two.

\section{Proof of Lemma 2}
For simplicity, we denote $\mathbb{E}_{v\sim\mathcal{N}(0, C)}$ by $\mathbb{E}$.
The first statement can be proved by definition:
\begin{align*}
\|\nabla h_{\mu,C}(y) - \nabla h_{\mu,C}(x)\|_* 
&=  \|\nabla \mathbb{E}[h(y+\mu v)]-\nabla \mathbb{E}[h(x+\mu v)]\|_* \\
&=\|\mathbb{E}[\nabla h(y+\mu v)-\nabla h(x+\mu v)]\|_* \\
&\leq  \mathbb{E}[\|\nabla h(y+\mu v)-\nabla h(x+\mu v)\|_*] \\
&\leq  L\|y-x\|,
\end{align*}
where the first equation follows from \cref{nga}, the first inequality uses Jensen's inequality, and the last step is due to the $L$-smoothness of $h$.

The second statement can be derived from the smoothness assumption and the properties of Gaussian variables:
\[
\begin{split}
\|\nabla h_{\mu,C}(x) - \nabla h(x)\|_* 
\overset{(\ref{eq:gradient_gaussian_smoothing})}=& \left\|\mathbb{E}\left[\frac{h(x+\mu v)-h(x-\mu v)}{2\mu}C^{-1}v\right]-\nabla h(x)\right\|_* \\
\overset{(a)}=& \left\|\mathbb{E}\left[\frac{h(x+\mu v)-h(x-\mu v)}{2\mu}C^{-1}v\right] -C^{-1}\mathbb{E}[vv^\top]\nabla h(x)\right\|_* \\
= &\left\|\mathbb{E}\left[\left( \frac{h(x+\mu v)-h(x-\mu v)}{2\mu}-v^\top\nabla h(x)  \right)C^{-1}v\right]\right\|_* \\
\rrr{\le } &\mathbb{E}\left[\left|\frac{h(x+\mu v)-h(x-\mu v)}{2\mu}-v^\top \nabla h(x)\right| \|C^{-1} v\|_*\right] \\
\overset{(b)}\leq& \frac{L\mu}{2}\mathbb{E}[\| v\|^2\|C^{-1}v\|_*],
\end{split}
\]
where $(a)$ uses the properties of the Gaussian distribution, and $(b)$ is due to 
\begin{equation}\label{eq:another-bound-of-smoothness}
|h(x+\mu v) - h(x-\mu v) - 2\langle \nabla h(x),\mu v \rangle| \leq L\|\mu v\|^2,
\end{equation}
which follows from \cref{func_bound}.

The third statement, again, can be obtained using the smoothness assumption. 
\begin{equation*}
\begin{split}
|h_{\mu, C}(x)-h(x)| = &|\mathbb{E}[h(x+\mu v)]-h(x)| \\
\overset{(a)}=& |\mathbb{E}[h(x+\mu v)-h(x) - \mu v^\top \nabla h(x)]| \\
\overset{(b)}\leq& \mathbb{E}[|h(x+\mu v)-h(x) - \mu v^\top \nabla h(x)|] \\
\overset{(\ref{func_bound})}\leq& \frac{\mu^2L}{2}\mathbb{E}\left[\|v\|^2\right],
\end{split}
\end{equation*}
where $(a)$ is due to $\mathbb{E}[v] = 0$, and $(b)$ uses Jensen's inequality.

\section{Solution to Problem (10)}
\label{append:why-sensitiveness-matters}
Let $u_1 \sim \mathcal{N}(0,I_n), u_2 \sim \mathcal{N}(0,I_m)$ be two independent vectors. Let $v = \sqrt{1-\alpha}u_1 + \sqrt{\alpha}Q u_2$. It is then clear that $v \sim \mathcal{N}(0,C)$. Therefore, we can write the objective function as
\[
    \begin{split}
         \mathbb{E}[\|v\|^2] 
        &= \mathbb{E}[v^\top H v] \\
        &=  \mathbb{E}[(\sqrt{1-\alpha}u_1 + \sqrt{\alpha}Q u_2)^\top H (\sqrt{1-\alpha}u_1 + \sqrt{\alpha}Q u_2)] \\
        &=  (1-\alpha)\mathbb{E}[u_1^\top H u_1] + \alpha\mathbb{E}[ u_2^\top Q^\top H Q u_2]
    \end{split}
\]
where the first equation follows from the quadratic function assumption, and the last uses the fact $\mathbb{E}[u_1] = 0$ and $\mathbb{E}[u_2] = 0$. The first term on the rightmost side of the above does not involve $Q$, so 
\[
    \begin{split}
        \arg\min \mathbb{E}[\|v\|^2]  
        & = \arg\min \mathbb{E}[ u_2^\top Q^\top H Q u_2] \\
        & = \arg\min \mathbb{E}[  \Tr[Q^\top H Q u_2 u_2^\top]]\\
        & = \arg\min \Tr[Q^\top H Q \mathbb{E}[  u_2 u_2^\top]]\\
        & = \arg\min \Tr[Q^\top H Q] 
    \end{split} 
\]
where the last step uses the fact $\mathbb{E}[  u_2 u_2^\top] = I_m$.
The solution to the above minimization problem, under the constraint $Q^\top Q = I_m$, is well known, and is given by the eigenvectors corresponding to the $m$ smallest eigenvalues of $H$.
That is, $Q$ corresponds to the $m$ principal components of $H^{-1}$.

\section{Proof of Theorem 1}
\label{append:proof-of-convergence-theorem}

We suppose that the number of rounds $R$  is sufficiently large such that the step-size $\eta$ defined in \cref{eq:constant-step-size-setting} satisfies the following conditions: 
\begin{align}
    & \eta \le \frac{1}{8LK\sqrt{n+4}}, \label{eq:condition-on-step-size-1} \\
    & \eta \le \frac{1}{LK^{1.5}\sqrt{10M}},\label{eq:condition-on-step-size-2} \\
    & \eta \le \frac{1-\alpha}{\rrr{3} LK},\label{eq:condition-on-step-size-3} \\
    & \eta \le \frac{M(1-\alpha)}{\rrr{384}L(n+4)},\label{eq:condition-on-step-size-4} \\
    & \eta \le \frac{1}{MLK^2},\label{eq:condition-on-step-size-5} \\
    & \eta \le \frac{1}{\rrr{15}LK},\label{eq:condition-on-step-size-6} \\
    & \eta \le \frac{M}{\rrr{8}L(1-\alpha)}.\label{eq:condition-on-step-size-7}
\end{align}

For simplicity, denote $f^i$ as the local objective function of client $i$ and $\tilde{f}_r^i$ as its Gaussian smoothing at round $r$, i.e., 
\[
    f^i(x) = \mathbb{E}_{\xi \sim \mathcal{D}_i}[F(x;\xi)],
\]
and
\[
    \tilde{f}_r^i(x) = \mathbb{E}_{v\sim \mathcal{N}(0,C_r)}[f^i(x+\mu v)].
\]  

We first state some properties of Gaussian smoothing under the $\ell_2$ norm assumption.
\begin{lemma}\label{lemma:gaussian-smoothing-gradient-estimate-bound}
    Let $h:\mathbb{R}^n \to \mathbb{R}$ be an $L$-smooth function w.r.t. the $\ell_2$ norm $\|\cdot\| = \|\cdot\| = \|\cdot\|_2$.
    Let $g$ be a gradient estimation to its Gaussian smoothing $h_{\mu,C}$ given in \cref{eq:unbiased-gradient-estimation-of-gaussian-smoothing} and assume the covariance matrix $C$ is constrained in the form of \cref{eq:covariance-matrix-model}. Then we have
    \[
    \mathbb{E}[\|g\|^2] \le \frac{\mu^2}{2}L^2(n+6)^3 + 2(n+4)\|\nabla h(x)\|^2
    \] 
    for all $x\in \mathbb{R}^n$.
    In addition, the bounds in \cref{eq:gaussian-smoothing-bound-on-gradient,eq:gaussian-smoothing-bound-on-objective} can be tightened as
    \begin{equation}
        \label{eq:gaussian-smoothing-bound-on-gradient-l2}
        \|\nabla h_{\mu,C}(x) - \nabla h(x)\| \le \frac{L \mu}{2\sqrt{1-\alpha}}(n+3)^{1.5},
    \end{equation}
    and
    \begin{equation}
        \label{eq:gaussian-smoothing-bound-on-objective-l2}
        |h_{\mu,C}(x) - h(x)| \le  \frac{nL\mu^2}{2}.
    \end{equation}
\end{lemma}

In the following we show that the client drifts are bounded.  

\begin{lemma}\label{lemma:bounds-of-client-drift}
    Under the same assumptions as in \Cref{theorem:convergence}, we have
    \begin{equation}\label{eq:bound-of-client-shift}
    \begin{split}
        \sum_{k=0}^{K-1} \mathbb{E}\left[\|w_{r,k}^i - x_r\|^2\right] \le 2 \eta^2 K^3 (\psi + 8(n+4) \|\nabla f(x_r) \|^2), 
    \end{split}
    \end{equation}
    and
    \begin{equation}\label{eq:bound-of-local-gradient-shift}
        \sum_{k=0}^{K-1} \mathbb{E}\left[\|g_{r,k}^i\|^2\right] \le 2K (\psi + 8(n+4) \|\nabla f(x_r) \|^2),
    \end{equation}
    where 
    \begin{equation}\label{eq:psi}
    \psi = \frac{1}{2}\mu^2L^2(n+6)^3 + 2(n+4)(4\sigma_g^2+\sigma_l^2),
    \end{equation}
    \rrr{and the expectation is conditioned on the history available at the beginning of round $r$\footnote{\rrr{Hence, $x_r$ and $C_r$ are fixed in the analysis. The same notation is also used in \Cref{lemma:sufficient-descent,lemma:bound-of-descent-step}.}}.}
\end{lemma}

The following bound shows that the global update of the solution gives a sufficient descent direction.    
\begin{lemma}\label{lemma:sufficient-descent}
    Under the same assumptions as in \Cref{theorem:convergence}, we have
\begin{equation*}
\begin{split}
    \mathbb{E} &\left[ \left<\nabla f(x_r), x_{r+1}-x_r \right> \right]
    \le -\frac{\eta K}{4}(1-\alpha)\left\| \nabla f(x_r) \right\|^2 \\
    & - \frac{\eta K}{2}(1-\alpha)\mathbb{E}\left[ \left\|  \frac{1}{NK}\sum_{i=1}^N \sum_{k=0}^{K-1} \nabla \tilde{f}_r^i(w_{r,k}^i) \right\|^2  \right]  + 2L^2 \eta^3 K^3\psi + \frac{\eta KL^2 \mu^2}{4(1-\alpha)}(n+3)^3,
\end{split}    
\end{equation*}
where $\psi$ is a constant given in \cref{eq:psi}.
\end{lemma}

\begin{lemma}\label{lemma:bound-of-descent-step}
With the same assumptions in \Cref{theorem:convergence}, we have
\[
    \begin{split}
    \mathbb{E}\left[ \|x_{r+1} - x_r\|^2 \right]  
    \le & \frac{\rrr{96}\eta^2K}{M} (n+4) \left\| \nabla f(x_r) \right\|^2 \\
    & + \frac{\rrr{15\eta^2 K^2} L^2 \mu^2 (n+3)^3}{2(1-\alpha)} + \frac{\rrr{12}\eta^2K}{M}\psi  + \rrr{15}\eta^2K^2 \frac{\sigma_g^2}{M} \\
    & + \rrr{3} \eta^2 \mathbb{E}\left[ \left\|  \frac{1}{N}\sum_{i=1}^N\sum_{k=0}^{K-1}\nabla \tilde{f}_r^i(w_{r,k}^i) \right\|^2  \right]
    \end{split}
\]
where $\psi$ is a constant given in \cref{eq:psi}.
\end{lemma}

\begin{proof}[Proof of \Cref{theorem:convergence}]
    Using the smoothness assumption, we have 
\begin{equation*}
    \mathbb{E}[f(x_{r+1})] \le f(x_r) + \mathbb{E}\left[ \left<\nabla f(x_r),x_{r+1}-x_r \right> \right] + \frac{L}{2}\mathbb{E}\left[ \left\| x_{r+1}-x_r \right\|^2 \right].
\end{equation*}
The right-hand side can be bounded using \Cref{lemma:bound-of-descent-step,lemma:sufficient-descent}, which yields
\begin{equation}\label{eq:3T02L46CSJ}
    \begin{split}
        \mathbb{E}[f(x_{r+1})] &\le f(x_r) 
        -\frac{\eta K}{4}(1-\alpha)\left\| \nabla f(x_r) \right\|^2  \\
        & - \frac{\eta K}{2}(1-\alpha)\mathbb{E}\left[ \left\|  \frac{1}{NK}\sum_{i=1}^N \sum_{k=0}^{K-1} \nabla \tilde{f}_r^i(w_{r,k}^i) \right\|^2  \right]  \\
        & + 2L^2 \eta^3 K^3\psi + \frac{\eta KL^2 \mu^2}{4(1-\alpha)}(n+3)^3  \\
        & + \frac{\rrr{48}L\eta^2K}{M} (n+4) \left\| \nabla f(x_r) \right\|^2  \\
        & + \frac{L}{2}\left(   \frac{\rrr{15\eta^2 K^2} L^2 \mu^2 (n+3)^3}{2(1-\alpha)} + \frac{\rrr{12}\eta^2K}{M}\psi  + \rrr{15}\eta^2K^2 \frac{\sigma_g^2}{M}  \right)  \\
        & + \frac{\rrr{3}L}{2}\eta^2 \mathbb{E}\left[ \left\|  \frac{1}{N}\sum_{i=1}^N\sum_{k=0}^{K-1}\nabla \tilde{f}_r^i(w_{r,k}^i) \right\|^2  \right]  \\
        =& f(x_r) 
        -\eta K\left(  \frac{1-\alpha}{4} - \frac{\rrr{48}L\eta}{M} (n+4)  \right) \left\| \nabla f(x_r) \right\|^2  \\
        & - \frac{\eta}{2}\left(\frac{1-\alpha}{K} - \rrr{3}L\eta \right) \mathbb{E}\left[ \left\|  \frac{1}{N}\sum_{i=1}^N\sum_{k=0}^{K-1}\nabla \tilde{f}_r^i(w_{r,k}^i) \right\|^2  \right]  \\
        & + 2L^2 \eta^3 K^3\psi + \frac{\eta KL^2 \mu^2}{4(1-\alpha)}(n+3)^3  \\
        & +  \frac{\rrr{15 \eta^2 K^2} L^3 \mu^2 (n+3)^3}{4(1-\alpha)} + \frac{\rrr{6} L\eta^2K}{M}\psi  + \eta^2K^2 \frac{\rrr{15} L\sigma_g^2}{2M}  \\
        \overset{(\ref{eq:condition-on-step-size-3},\ref{eq:condition-on-step-size-4})}\le& f(x_r) -\eta K \frac{1-\alpha}{8}  \left\| \nabla f(x_r) \right\|^2  \\
        & + \eta K \left(
        2L\eta\left(L \eta K^2 + \frac{\rrr{3}}{M}\right)\psi + \frac{L^2 \mu^2(n+3)^3}{4(1-\alpha)}(1+\rrr{15}L\eta K) + \eta K \frac{\rrr{15}L\sigma_g^2}{2M} \right) \\
        \overset{(\ref{eq:condition-on-step-size-5},\ref{eq:condition-on-step-size-6})}\le & f(x_r) -\eta K \frac{1-\alpha}{8}  \left\| \nabla f(x_r) \right\|^2   \\
        &+ \eta K \underbrace{\left( \frac{\rrr{8}}{M}L\eta\psi + \frac{L^2 \mu^2(n+3)^3}{2(1-\alpha)} + \eta K \frac{\rrr{15}L\sigma_g^2}{2M} \right)}_{\mathfrak{E}}.
    \end{split}
\end{equation} 

Now substituting $\psi$ with \cref{eq:psi} and using the condition $K \le n$, we bound the term $\mathfrak{E}$ as

\[
    \begin{split}
        \mathfrak{E} = & \rrr{8} L\eta \left(  \frac{1}{2M}\mu^2L^2(n+6)^3 + \frac{2}{M}(n+4)(4\sigma_g^2+\sigma_l^2)  \right) + \frac{L^2 \mu^2(n+3)^3}{2(1-\alpha)} + \eta K \frac{\rrr{15} L\sigma_g^2}{2M} \\
        \le & \frac{\rrr{4}}{M}\eta \mu^2L^3(n+6)^3 + \frac{\rrr{16}}{M}L\eta(n+4)(4\sigma_g^2+\sigma_l^2)  + \frac{L^2 \mu^2(n+3)^3}{2(1-\alpha)}  + \eta n \frac{\rrr{15} L\sigma_g^2}{2M} \\
        \le & \frac{\rrr{4}}{M}\eta \mu^2L^3(n+6)^3 + \frac{L^2 \mu^2(n+3)^3}{2(1-\alpha)} + \frac{\rrr{18}}{M}L\eta(n+4)(4\sigma_g^2+\sigma_l^2) \\
        \le & \mu^2 L^2(n+6)^3\left(  \frac{\rrr{4} \eta L}{M} + \frac{1}{2(1-\alpha)} \right) + \frac{\rrr{18}}{M}L\eta(n+4)(4\sigma_g^2+\sigma_l^2) \\
        \overset{\cref{eq:condition-on-step-size-7}}\le & \frac{\mu^2 L^2(n+6)^3}{1-\alpha} + \frac{\rrr{72}}{M}L\eta(n+4)(\sigma_g^2+\sigma_l^2).
    \end{split}
\] 
Plugging the above into \cref{eq:3T02L46CSJ}, we have
\[
    \begin{split}
    \mathbb{E}[f(x_{r+1})] 
    &\le f(x_r) -\eta K \frac{1-\alpha}{8}  \left\| \nabla f(x_r) \right\|^2
    \\
    &\quad + \eta K \left(   \frac{\mu^2 L^2(n+6)^3}{1-\alpha} + \frac{\rrr{72}}{M}L\eta(n+4)(\sigma_g^2+\sigma_l^2) \right).
    \end{split}
\]
Summing over $r=0,\dots,R-1$, taking total expectation, and using \Cref{assumption:objective-function-lower-bound}, we have
\[
\begin{split}
    \frac{1}{R}\sum_{r=0}^{R-1}\mathbb{E}\left[  \left\| \nabla f(x_r) \right\|^2 \right]
    &\le \frac{8\mu^2 L^2(n+6)^3}{(1-\alpha)^2} \\
    &\quad + \frac{8}{1-\alpha}\left(\frac{f(x_0)-f_*}{R \eta K} + \frac{\rrr{72}}{M}L\eta(n+4)(\sigma_g^2+\sigma_l^2)\right) \\
    &\le \frac{\rrr{192}}{1-\alpha}\sqrt{\frac{(f(x_0)-f_*)(n+4)(\sigma_g^2+\sigma_l^2)\rrr{L}}{RKM}} \\
    &\quad + \frac{8\mu^2 L^2(n+6)^3}{(1-\alpha)^2}
\end{split}
\]
where in the second inequality, we use the step-size setting \cref{eq:constant-step-size-setting}.
\end{proof}

\section{Proof of Lemma 3}
The proof follows \cite[Theorem 4]{nesterov2017random} with slight modifications due to the non-isotropic covariance matrix:
\begin{equation*}
    \begin{split}
        \|g\|^2  \overset{\cref{eq:unbiased-gradient-estimation-of-gaussian-smoothing}} = & \left\| \frac{h(x + \mu v) - h(x - \mu v)}{2 \mu} v\right\|^2 \\ 
        = & \frac{1}{4\mu^2}\left\|
        (h(x + \mu v) - h(x - \mu v) - 2 \langle \nabla h(x), \mu v\rangle) v + 2 \langle \nabla h(x), \mu v\rangle v \right\|^2 \\ 
        \le & \frac{1}{2\mu^2}\| (h(x + \mu v) - h(x - \mu v) - 2 \langle \nabla h(x), \mu v\rangle) v \|^2  + \frac{1}{2\mu^2} \|\langle 2\nabla h(x), \mu v\rangle v \|^2 \\ 
        = & \frac{1}{2\mu^2}|h(x + \mu v) - h(x - \mu v) - 2 \langle \nabla h(x), \mu v\rangle|^2\|  v \|^2  + 2 \langle \nabla h(x), v\rangle^2 \| v \|^2 \\ 
        \overset{\cref{eq:another-bound-of-smoothness}}\le & \frac{1}{2\mu^2}(L\|\mu v\|^2)^2  \|v\|^2 + 2 \langle \nabla h(x), v\rangle^2 \| v \|^2 \\ 
        = & \frac{\mu^2L^2}{2}  \|v\|^6 + 2 \langle \nabla h(x), v\rangle^2 \| v \|^2
    \end{split}
\end{equation*}
where the first inequality uses the fact 
\begin{equation}\label{eq:bound-of-sum-norm}
    \left\| \sum_{i=1}^k a_i \right\|^2 \le k \sum_{i=1}^k \|a_i\|^2.
\end{equation} 
Now let $v = \sqrt{C} u$ where $\sqrt{C}$ denotes the 0.5th power of $C$. We have
\begin{equation*}
    \begin{split}
        \|g\|^2 & \le \frac{\mu^2L^2}{2}  \|\sqrt{C} u\|^6 + 2 \langle \sqrt{C} \nabla h(x), u\rangle^2\| \sqrt{C} u \|^2 \\
        & = \frac{\mu^2L^2}{2} (u^\top C u)^3 + 2  \langle \sqrt{C} \nabla h(x), u\rangle^2 (u^\top C u).
    \end{split}
\end{equation*}
Recall that the covariance matrix in the form of \cref{eq:covariance-matrix-model} can be bounded as $(1-\alpha)I_n \preceq C \preceq I_n$. We have $u^\top C u \le \|u\|^2$ and therefore
\begin{equation*}
        \|g\|^2 \le \frac{\mu^2L^2}{2} \|u\|^6 + 2  \langle \sqrt{C} \nabla h(x), u\rangle^2 \|u\|^2.
\end{equation*}
Taking expectation gives:
\begin{equation*}
    \begin{split}
        \mathbb{E}[\|g\|^2] & = \mathbb{E}_{u \sim \mathcal{N}(0,I_n)}[\|g\|^2] \\
        & \le \frac{\mu^2L^2}{2} \mathbb{E}[\|u\|^6] + 2  \mathbb{E}[\langle \sqrt{C} \nabla h(x), u\rangle^2 \|u\|^2] \\
        & \le \frac{\mu^2L^2}{2} (n+6)^3 + 2 (n+4)\|\sqrt{C} \nabla h(x)\|^2
    \end{split}
\end{equation*}
where the last inequality follows from \cite[equations (17) and (32)]{nesterov2017random}. Using the fact $\|\sqrt{C} \nabla h(x)\|^2 \le \|\rrr{\nabla} h(x)\|^2$ derived from $C \preceq I_n$, we reach the first statement.

Under the $\ell_2$ norm assumption, we have $\|v\|^2 = \|\sqrt{C}u\|^2 \le \|u\|^2$ and $\|C^{-1}v\| = \|C^{-\frac{1}{2}}u\| \le \frac{1}{\sqrt{1-\alpha}}\|u\|$. Plugging these bounds into \cref{eq:gaussian-smoothing-bound-on-gradient,eq:gaussian-smoothing-bound-on-objective} and using \cite[Lemma 1]{nesterov2017random}, we can obtain \cref{eq:gaussian-smoothing-bound-on-gradient-l2,eq:gaussian-smoothing-bound-on-objective-l2}, respectively.

\section{Proof of Lemma 4}
Applying \Cref{lemma:gaussian-smoothing-gradient-estimate-bound} to $g_{r,k}^i$ (via replacing $h(x)$ with $F(w_{r,k}^i;\xi_{r,k}^i)$), we have
\begin{equation}\label{eq:JCPVL5HFQF}
\begin{split}
    \mathbb{E}\left[\left\| g_{r,k}^i\right\|^2\right] \le &\frac{\mu^2}{2}L^2(n+6)^3 + 2(n+4) \mathbb{E}\left[\|\nabla F(w_{r,k}^i;\xi_{r,k}^i)\|^2\right].
\end{split}
\end{equation}
The stochastic gradient norm on the right-hand side can be bounded as
\[
\begin{split}
    \mathbb{E}\left[\left\|\nabla F(w_{r,k}^i; \xi_{r,k}^i)\right\|^2\right]
    \overset{(a)}\le & \mathbb{E}\left[\left\|\nabla f^i(w_{r,k}^i)\right\|^2\right] + \sigma_l^2 \\
    = & \mathbb{E}\left[\left\|\nabla f^i(w_{r,k}^i) - \nabla f^i(x_r) + \nabla f^i(x_r)\right\|^2\right] + \sigma_l^2 \\
    \overset{\cref{eq:bound-of-sum-norm}}\le & 2\left\|\nabla f^i(x_r)\right\|^2 + 2\mathbb{E}\left[\left\|\nabla f^i(w_{r,k}^i) - \nabla f^i(x_r)\right\|^2\right] + \sigma_l^2 \\
    \overset{(b)}\le & 2\left\|\nabla f^i(x_r)\right\|^2 + 2L^2\mathbb{E}\left[\left\|w_{r,k}^i - x_r\right\|^2\right] + \sigma_l^2 \\
    =&  2\left\|\nabla f^i(x_r) - \nabla f(x_r) + \nabla f(x_r)\right\|^2  \\
    &\quad +  2L^2\mathbb{E}\left[\left\|w_{r,k}^i - x_r\right\|^2\right] + \sigma_l^2 \\
    \overset{\cref{eq:bound-of-sum-norm}} \le & 4\left\|\nabla f^i(x_r) - \nabla f(x_r)\right\|^2 + 4\|\nabla f(x_r)\|^2 \\
    &\quad + 2L^2\mathbb{E}\left[\left\|w_{r,k}^i - x_r\right\|^2\right] + \sigma_l^2 \\
    \overset{(c)}\le & 4\sigma_g^2 + 4\|\nabla f(x_r)\|^2 \\
    &\quad + 2L^2\mathbb{E}\left[\left\|w_{r,k}^i - x_r\right\|^2\right] + \sigma_l^2
\end{split}     
\]
where $(a)$ follows from \Cref{assumption:local-gradient-variance-bound}, $(b)$ is due to \Cref{assumption:objective-smoothness}, and $(c)$ is due to \Cref{assumption:global-gradient-bound}.
Plugging the above into \cref{eq:JCPVL5HFQF} and summing over $k = 0, \dots, K-1$, one obtains
\begin{equation}\label{eq:GLZ4NDQGV3}
    \begin{aligned}[b]
    \sum_{k=0}^{K-1}\mathbb{E}\left[\left\| g_{r,k}^i\right\|^2\right] 
    \le& 8K(n+4)\|\nabla f(x_r)\|^2  + K\left(\frac{\mu^2}{2}L^2(n+6)^3 + 2(n+4)(4\sigma_g^2 + \sigma_l^2)\right) \\
    & + 4L^2(n+4) \sum_{k=0}^{K-1}\mathbb{E}\left[\left\|w_{r,k}^i - x_r\right\|^2\right] \\
    = & K(\psi + 8(n+4)\|\nabla f(x_r)\|^2)  + 4L^2(n+4) \sum_{k=0}^{K-1}\mathbb{E}\left[\left\|w_{r,k}^i - x_r\right\|^2\right].
    \end{aligned}
\end{equation}

On the other hand, by construction and the fact $w_{r,0}^i = x_r$, we have
\begin{equation*}
\begin{split}
    \sum_{k=0}^{K-1} \mathbb{E}\left[\|w_{r,k}^i - x_r\|^2\right] 
    =& \sum_{k=0}^{K-1} \mathbb{E}\left[\left\|\sum_{j=0}^{k-1}w_{r,j+1}^i - w_{r,j}^i\right\|^2\right] \\
    =& \sum_{k=0}^{K-1} \mathbb{E}\left[\left\|\sum_{j=0}^{k-1} \eta g_{r,j}^i\right\|^2\right] \\
    \overset{\cref{eq:bound-of-sum-norm}}\le & \eta^2\sum_{k=0}^{K-1}  k \sum_{j=0}^{k-1} \mathbb{E}\left[\left\| g_{r,j}^i\right\|^2\right] \\
    \le& \eta^2\sum_{k=0}^{K-1}  K \sum_{j=0}^{K-1} \mathbb{E}\left[\left\| g_{r,j}^i\right\|^2\right] \\
    =& \eta^2 K^2 \sum_{k=0}^{K-1} \mathbb{E}\left[\left\| g_{r,k}^i\right\|^2\right] \\
    \overset{\cref{eq:GLZ4NDQGV3}}\le & \eta^2 K^3 (\psi + 8(n+4)\|\nabla f(x_r)\|^2) \\
    &+ 4 \eta^2 K^2 L^2(n+4) \sum_{k=0}^{K-1}\mathbb{E}\left[\left\|w_{r,k}^i - x_r\right\|^2\right] \\
    \overset{\cref{eq:condition-on-step-size-1}}\le &\eta^2 K^3 (\psi + 8(n+4)\|\nabla f(x_r)\|^2) + \frac{1}{2} \sum_{k=0}^{K-1}\mathbb{E}\left[\left\|
        w_{r,k}^i - x_r
    \right\|^2\right].  
\end{split}
\end{equation*}
Rearranging the terms, one obtains \cref{eq:bound-of-client-shift}. The bound in \cref{eq:bound-of-local-gradient-shift} can be obtained by putting \cref{eq:bound-of-client-shift} back into \cref{eq:GLZ4NDQGV3}.

\section{Proof of Lemma 5}
By construction, we write the per-round descent step as
\[
    \begin{split}
    \mathbb{E}[x_{r+1}-x_r] & = -\mathbb{E}\left[ \frac{1}{M}\sum_{i\in \mathcal{W}_r}\sum_{k=0}^{K-1} \eta g_{r,k}^i  \right]  \\
    & = -\eta\mathbb{E}\left[ \frac{1}{M}\sum_{i\in \mathcal{W}_r}\sum_{k=0}^{K-1} C_r \nabla \tilde{f}_r^i(w_{r,k}^i)  \right]  \\
    & = -\eta C_r \frac{1}{N}\sum_{i=1}^N \sum_{k=0}^{K-1} \mathbb{E}\left[\nabla \tilde{f}_r^i(w_{r,k}^i)  \right]
    \end{split}
\]
where the second equation follows from \Cref{lemma:differentiability} and the third equation is due to the uniform sampling of the client set.
It follows that 
\begin{equation}\label{eq:2W3NTIC6SA}
\begin{split}
\mathbb{E}& \left[ \left<\nabla f(x_r), x_{r+1}-x_r \right> \right] \\
&\quad = -\eta K \mathbb{E}\left[ \left<\nabla f(x_r),  C_r \frac{1}{NK}\sum_{i=1}^N \sum_{k=0}^{K-1} \nabla \tilde{f}_r^i(w_{r,k}^i)\right> \right]  \\
&\quad = -\eta K \mathbb{E}\left[ \left< \sqrt{C_r} \nabla f(x_r), \sqrt{C_r} \frac{1}{NK}\sum_{i=1}^N \sum_{k=0}^{K-1} \nabla \tilde{f}_r^i(w_{r,k}^i)
\right> \right]  \\
&\quad = -\frac{\eta K}{2}\left\| \sqrt{C_r}\nabla f(x_r) \right\|^2 - \frac{\eta K}{2}\mathbb{E}\left[ \left\| \sqrt{C_r} \frac{1}{NK}\sum_{i=1}^N \sum_{k=0}^{K-1} \nabla \tilde{f}_r^i(w_{r,k}^i) \right\|^2  \right]   \\
&\qquad + \frac{\eta K}{2} \nonumber\mathbb{E}\left[ \left\|\sqrt{C_r}\nabla f(x_r) - \sqrt{C_r} \frac{1}{NK}\sum_{i=1}^N\sum_{k=0}^{K-1}\nabla \tilde{f}_r^i(w_{r,k}^i)\right\|^2  \right] \\
& \quad \le -\frac{\eta K}{2}(1-\alpha)\left\| \nabla f(x_r) \right\|^2  - \frac{\eta K}{2}(1-\alpha)\mathbb{E}\left[ \left\|  \frac{1}{NK}\sum_{i=1}^N \sum_{k=0}^{K-1} \nabla \tilde{f}_r^i(w_{r,k}^i) \right\|^2  \right]  \\
& \qquad + \frac{\eta K}{2} \underbrace{\mathbb{E}\left[ \left\| \nabla f(x_r) - \frac{1}{NK}\sum_{i=1}^N\sum_{k=0}^{K-1}\nabla \tilde{f}_r^i(w_{r,k}^i) \right\|^2  \right]}_{\mathfrak{A}}, %
\end{split}
\end{equation}

where the third equation uses the identity $\left<a,b \right> = \frac{1}{2}\left( \|a\|^2 + \|b\|^2-\|a-b\|^2 \right)$ and the inequality uses the bound on the covariance matrix $(1-\alpha)I_n \preceq C_r \preceq I_n$.

Using the fact $f(x) = \frac{1}{N}\sum_{i=1}^N f^i(x)$, we can bound $\mathfrak{A}$ in \cref{eq:2W3NTIC6SA} as
\[
    \begin{split}
    \mathfrak{A} 
    = & \mathbb{E}\left[ \left\| \frac{1}{NK}\sum_{i=1}^N\sum_{k=0}^{K-1} \nabla f^i(x_r) - \nabla \tilde{f}_r^i (w_{r,k}^i) \right\|^2  \right]      \\
    \le & \frac{1}{NK}\sum_{i=1}^N\sum_{k=0}^{K-1}\mathbb{E}\left[ \left\| \nabla f^i(x_r) - \nabla \tilde{f}_r^i (w_{r,k}^i) \right\|^2  \right]   \\
    = & \frac{1}{NK} \sum_{i=1}^N\sum_{k=0}^{K-1}\mathbb{E}\left[ \left\| \nabla f^i(x_r) - \nabla f^i(w_{r,k}^i) + \nabla f^i(w_{r,k}^i) - \nabla \tilde{f}_r^i (w_{r,k}^i)
    \right\|^2  \right]     \\
    \overset{\cref{eq:bound-of-sum-norm}}\le & \frac{2}{NK}\sum_{i=1}^N\sum_{k=0}^{K-1}\mathbb{E}\left[ \left\| \nabla f^i(x_r) - \nabla f^i(w_{r,k}^i)  \right\|^2  \right] \\
    &\quad + \frac{2}{NK}\sum_{i=1}^N\sum_{k=0}^{K-1}\mathbb{E}\left[ \left\| \nabla f^i(w_{r,k}^i) - \nabla \tilde{f}_r^i (w_{r,k}^i) \right\|^2  \right] \\
    \overset{(a)}\le & \frac{2L^2}{NK}\sum_{i=1}^N\sum_{k=0}^{K-1}\mathbb{E}\left[ \left\| x_r - w_{r,k}^i  \right\|^2  \right]  \\
    &\quad + \frac{2}{NK}\sum_{i=1}^N\sum_{k=0}^{K-1}\mathbb{E}\left[ \left\| \nabla f^i(w_{r,k}^i) - \nabla \tilde{f}_r^i (w_{r,k}^i) \right\|^2  \right] \\
    \overset{\cref{eq:gaussian-smoothing-bound-on-gradient-l2}}\le & \frac{2L^2}{NK}\sum_{i=1}^N\sum_{k=0}^{K-1}\mathbb{E}\left[ \left\| x_r - w_{r,k}^i  \right\|^2  \right] 
    + \frac{L^2 \mu^2}{2(1-\alpha)}(n+3)^3 \\
    \overset{\cref{eq:bound-of-client-shift}}\le & 4L^2 \eta^2 K^2(\psi + 8(n+4)\|\nabla f(x_r)\|^2)  + \frac{L^2 \mu^2}{2(1-\alpha)}(n+3)^3,
    \end{split}
\]
where $(a)$ uses the smoothness assumption.
Plugging the above into \cref{eq:2W3NTIC6SA}, we have
\begin{equation*}
\begin{split}
    \mathbb{E}& \left[ \left<\nabla f(x_r), x_{r+1}-x_r \right> \right] \\
    &\quad \le -\frac{\eta K}{2}(1-\alpha)\left\| \nabla f(x_r) \right\|^2 - \frac{\eta K}{2}(1-\alpha)\mathbb{E}\left[ \left\|  \frac{1}{NK}\sum_{i=1}^N \sum_{k=0}^{K-1} \nabla \tilde{f}_r^i(w_{r,k}^i) \right\|^2  \right] \\
    &\qquad+ 2L^2 \eta^3 K^3(\psi + 8(n+4)\|\nabla f(x_r)\|^2) + \frac{\eta KL^2 \mu^2}{4(1-\alpha)}(n+3)^3 \\
    & \quad=  -\frac{\eta K}{2}(1-\alpha-32(n+4)\eta^2K^2L^2)\left\| \nabla f(x_r) \right\|^2 \\ 
    & \qquad - \frac{\eta K}{2}(1-\alpha)\mathbb{E}\left[ \left\|  \frac{1}{NK}\sum_{i=1}^N \sum_{k=0}^{K-1} \nabla \tilde{f}_r^i(w_{r,k}^i) \right\|^2  \right] \\
    & \qquad + 2L^2 \eta^3 K^3\psi + \frac{\eta KL^2 \mu^2}{4(1-\alpha)}(n+3)^3 \\ 
    &\quad \overset{\rrr{(\ref{eq:condition-on-step-size-4},\ref{eq:condition-on-step-size-5})}}\le  -\frac{\eta K}{4}(1-\alpha)\left\| \nabla f(x_r) \right\|^2  - \frac{\eta K}{2}(1-\alpha)\mathbb{E}\left[ \left\|  \frac{1}{NK}\sum_{i=1}^N \sum_{k=0}^{K-1} \nabla \tilde{f}_r^i(w_{r,k}^i) \right\|^2  \right] \\
    &\qquad + 2L^2 \eta^3 K^3\psi + \frac{\eta KL^2 \mu^2}{4(1-\alpha)}(n+3)^3. 
\end{split}
\end{equation*}
The proof then completes.

\section{Proof of Lemma 6}
By construction, we have
\[
\begin{split}
&x_{r+1} - x_r =  \rrr{-}\eta \frac{1}{M}\sum_{i\in \mathcal{W}_r}\sum_{k=0}^{K-1} g_{r,k}^i \\
&\quad=  \rrr{-}\eta  \underbrace{\frac{1}{M}\sum_{i\in \mathcal{W}_r}\sum_{k=0}^{K-1} \left(g_{r,k}^i - \rrr{C_r} \nabla \tilde{f}_r^i(w_{r,k}^i)\right)}_{\mathfrak{B}}\\
&\qquad\rrr{-}  \eta \rrr{C_r} \underbrace{\left(  \frac{1}{M}\sum_{i\in \mathcal{W}_r}\sum_{k=0}^{K-1}   \nabla \tilde{f}_r^i(w_{r,k}^i) - \frac{1}{N}\sum_{i=1}^N\sum_{k=0}^{K-1} \nabla \tilde{f}_r^i(w_{r,k}^i)  \right)}_{\mathfrak{C}}\\
&\qquad\rrr{-}  \eta \frac{1}{N}\sum_{i=1}^N\sum_{k=0}^{K-1} \rrr{C_r} \nabla \tilde{f}_r^i(w_{r,k}^i).
\end{split}
\]
\rrr{Since $C_r \preceq I$, we have}
\begin{equation}\label{eq:8HUG0DYN3M}
\begin{split}
    \mathbb{E}\left[ \| x_{r+1} - x_r\|^2  \right]   
    &\le  \rrr{3} \eta^2 \mathbb{E}\left[ \|\mathfrak{B}\|^2  \right] 
    + \rrr{3} \eta^2 \mathbb{E}\left[ \|\mathfrak{C}\|^2  \right] \\
    &\quad + \rrr{3} \eta^2 \mathbb{E}\left[ \left\|  \frac{1}{N}\sum_{i=1}^N\sum_{k=0}^{K-1}\nabla \tilde{f}_r^i(w_{r,k}^i) \right\|^2  \right].
\end{split}
\end{equation}

The term $\mathfrak{B}$ can be bounded as
\begin{equation}\label{eq:OM7MI7SVA9}
    \begin{aligned}[b]
    \mathbb{E}\left[ \|\mathfrak{B}\|^2  \right] 
    & \overset{(a)}= \frac{1}{M^2} \mathbb{E}\left[  \sum_{i\in \mathcal{W}_r}\sum_{k=0}^{K-1}\left\| g_{r,k}^i - \rrr{C_r} \nabla \tilde{f}_r^i(w_{r,k}^i) \right\|^2  \right] \\
    & \overset{(b)}\le \frac{1}{M^2} \mathbb{E}\left[  \sum_{i\in \mathcal{W}_r}\sum_{k=0}^{K-1}\left\| g_{r,k}^i \right\|^2  \right] \\
    & \overset{(c)}= \frac{1}{MN} \mathbb{E}\left[  \sum_{i=1}^N \sum_{k=0}^{K-1}\left\| g_{r,k}^i \right\|^2  \right] \\
    & \overset{\cref{eq:bound-of-local-gradient-shift}}\le \frac{2K}{M} \left( \psi + 8(n+4)\left\| \nabla f(x_r) \right\|^2  \right),
    \end{aligned}
\end{equation}
where $(a)$ uses the fact $\mathbb{E}[g_{r,k}^i] = \rrr{C_r} \nabla \tilde{f}_r^i(w_{r,k}^i) $, $(b)$ uses the fact $\mathbb{E}[\|a-\mathbb{E}[a]\|^2] \le \mathbb{E}[\|a\|^2]$, and $(c)$ is due to the uniform sampling of the clients.

The term $\mathfrak{C}$ can be written as 
\[
    \begin{split}
\mathfrak{C} 
= & \underbrace{\frac{1}{M}\sum_{i\in \mathcal{W}_r} \sum_{k=0}^{K-1} \left(\nabla \tilde{f}_r^i(w_{r,k}^i) - \nabla f^i(w_{r,k}^i)\right) }_{\mathfrak{D}_1}\\
\qquad\qquad & + \underbrace{\frac{1}{M}\sum_{i\in \mathcal{W}_r} \sum_{k=0}^{K-1} \left(\nabla f^i(w_{r,k}^i) - \nabla f^i(x_r)\right) }_{\mathfrak{D}_2}\\
\qquad\qquad & + \underbrace{\frac{1}{M}\sum_{i\in \mathcal{W}_r} \sum_{k=0}^{K-1} \nabla f^i(x_r) - \frac{1}{N}\sum_{i=1}^N \sum_{k=0}^{K-1} \nabla f^i(x_r) }_{\mathfrak{D}_3}\\
\qquad\qquad & + \underbrace{\frac{1}{N}\sum_{i=1}^N \sum_{k=0}^{K-1} \left(\nabla f^i(x_r) - \nabla f^i(w_{r,k}^i)\right) }_{\mathfrak{D}_4}\\
\qquad\qquad & + \underbrace{\frac{1}{N}\sum_{i=1}^N \sum_{k=0}^{K-1} \left(\nabla f^i(w_{r,k}^i) - \nabla \tilde{f}_r^i(w_{r,k}^i)\right) }_{\mathfrak{D}_5},
    \end{split}
\]
and hence, we have
\begin{equation} \label{eq:ELDEQYW8J2}
    \mathbb{E}\left[ \| \mathfrak{C} \|^2 \right] 
    \overset{\cref{eq:bound-of-sum-norm}}\le 5\mathbb{E}\left[ \|\mathfrak{D}_1\|^2 \right]  + 5\mathbb{E}\left[ \|\mathfrak{D}_2\|^2 \right] + 5\mathbb{E}\left[ \|\mathfrak{D}_3\|^2 \right] 
    + 5\mathbb{E}\left[ \|\mathfrak{D}_4\|^2 \right] + 5\mathbb{E}\left[ \|\mathfrak{D}_5\|^2 \right].
\end{equation}

Using \cref{eq:bound-of-sum-norm} and Jensen's inequality, we have 
\[
\begin{split}
\mathbb{E}\left[ \|\mathfrak{D}_1\|^2 \right] \le \frac{K}{M} \mathbb{E}\left[  \sum_{i\in \mathcal{W}_r} \sum_{k=0}^{K-1} \left\|\nabla \tilde{f}_r^i(w_{r,k}^i) - \nabla f^i(w_{r,k}^i)\right\|^2 \right] \overset{\cref{eq:gaussian-smoothing-bound-on-gradient-l2}}\le  \frac{K^2L^2\mu^2(n+3)^{3}}{4(1-\alpha)}.
\end{split}
\]
In addition, we obtain
\[
    \begin{split}
        \mathbb{E}\left[ \|\mathfrak{D}_2\|^2 \right] 
        & \overset{\cref{eq:bound-of-sum-norm}}\le \frac{K}{M} \mathbb{E}\left[  \sum_{i\in \mathcal{W}_r} \sum_{k=0}^{K-1} \left\|\nabla f^i(w_{r,k}^i) - \nabla f^i(x_r)\right\|^2 \right] \\
        & \overset{(a)}= \frac{K}{N}   \sum_{i=1}^N \sum_{k=0}^{K-1} \mathbb{E}\left[\left\|\nabla f^i(w_{r,k}^i) - \nabla f^i(x_r)\right\|^2 \right] \\
        & \overset{(b)}\le \frac{KL^2}{N}   \sum_{i=1}^N \sum_{k=0}^{K-1} \mathbb{E}\left[\left\|w_{r,k}^i - x_r\right\|^2 \right] \\
        & \overset{\cref{eq:bound-of-client-shift}}\le 2 \eta^2 K^4L^2 (\psi + 8(n+4) \|\nabla f(x_r) \|^2)
    \end{split}
\]
where $(a)$ is due to the uniform sampling of clients and $(b)$ uses the smoothness assumption.

We can derive in the same way the bounds for $\mathfrak{D}_4$ and $\mathfrak{D}_5$:
\[
\mathbb{E}\left[ \|\mathfrak{D}_4\|^2 \right] \le  2 \eta^2 K^4L^2 (\psi + 8(n+4) \|\nabla f(x_r) \|^2)
\]
and
\[
\mathbb{E}\left[ \|\mathfrak{D}_5\|^2 \right] \le \frac{K^2L^2\mu^2(n+3)^{3}}{4(1-\alpha)}.
\]

The term $\mathfrak{D}_3$ can be bounded as
\[
    \begin{split}
    \mathbb{E}\left[ \|\mathfrak{D}_3\|^2 \right] 
    & \overset{\cref{eq:bound-of-sum-norm}}\le \sum_{k=0}^{K-1} K \mathbb{E}\left[ \left\| \frac{1}{M}\sum_{i\in \mathcal{W}_r} \nabla f^i(x_r) - \nabla f(x_r) \right\|^2  \right] \\
    & = K^2 \mathbb{E}\left[  \frac{1}{M^2}\sum_{i\in \mathcal{W}_r} \left\| \nabla f^i(x_r) - \rrr\nabla f(x_r)\right\|^2  \right]\\
    & \le \frac{K^2\sigma_g^2}{M}
    \end{split}
\]
where the equation follows from the fact $\mathbb{E}[\nabla f^i(x_r)] = \rrr\nabla f(x_r)$ (due to the uniform sampling of clients), and the second inequality uses \Cref{assumption:global-gradient-bound}.

Putting the above bounds on $\mathfrak{D}_1$ to $\mathfrak{D}_5$ into \cref{eq:ELDEQYW8J2} gives
\[
\begin{split}
    \mathbb{E}\left[ \|\mathfrak{C}\|^2 \right] 
    &\le  K^2 \frac{5L^2 \mu^2 (n+3)^3}{2(1-\alpha)} + 20L^2\eta^2K^4\psi \\
    &\quad + 160L^2\eta^2K^4(n+4)\left\| \nabla f(x_r) \right\|^2 + 5K^2 \frac{\sigma_g^2}{M}.
\end{split}
\]
Plugging the above and \cref{eq:OM7MI7SVA9} into \cref{eq:8HUG0DYN3M}, we obtain
\[
    \begin{split}
    \mathbb{E}\left[ \|x_{r+1} - x_r\|^2 \right] \le & \rrr{3} \Biggl\{ \frac{2K\eta^2}{M} \left( \psi + 8(n+4)\left\| \nabla f(x_r) \right\|^2  \right)  \\
    &+ \eta^2 K^2 \frac{5L^2 \mu^2 (n+3)^3}{2(1-\alpha)} + 20L^2\eta^4K^4\psi \\
    & + 160L^2\eta^4K^4(n+4)\left\| \nabla f(x_r) \right\|^2 \\
    & + 5\eta^2K^2 \frac{\sigma_g^2}{M} + \eta^2 \mathbb{E}\left[ \left\|  \frac{1}{N}\sum_{i=1}^N\sum_{k=0}^{K-1}\nabla \tilde{f}_r^i(w_{r,k}^i) \right\|^2  \right] \Biggr\} \\
    = & \rrr{48}\eta^2K\left(\frac{1}{M} + 10L^2\eta^2K^3\right)(n+4) \left\| \nabla f(x_r) \right\|^2 \\
    & + \rrr{3}\eta^2 K^2 \frac{5L^2 \mu^2 (n+3)^3}{2(1-\alpha)} 
    + \rrr{6}\eta^2K\left(\frac{1}{M} + 10L^2\eta^2K^3\right)\psi \\
    & + \rrr{15}\eta^2K^2 \frac{\sigma_g^2}{M} 
    + \rrr{3}\eta^2 \mathbb{E}\left[ \left\|  \frac{1}{N}\sum_{i=1}^N\sum_{k=0}^{K-1}\nabla \tilde{f}_r^i(w_{r,k}^i) \right\|^2  \right]\\
    \le & \frac{\rrr{96}\eta^2K}{M} (n+4) \left\| \nabla f(x_r) \right\|^2 + \eta^2 K^2 \frac{\rrr{15} L^2 \mu^2 (n+3)^3}{2(1-\alpha)} \\
    & + \frac{\rrr{12}\eta^2K}{M}\psi + \rrr{15}\eta^2K^2 \frac{\sigma_g^2}{M} + \rrr{3}\eta^2 \mathbb{E}\left[ \left\|  \frac{1}{N}\sum_{i=1}^N\sum_{k=0}^{K-1}\nabla \tilde{f}_r^i(w_{r,k}^i) \right\|^2  \right]
    \end{split}
\]
where the last step uses the condition \cref{eq:condition-on-step-size-2}. The proof completes.

\begin{credits}
\subsubsection{Disclosure of Interests}
The authors have no competing interests to declare that are relevant to the content of this article.
\end{credits}

\bibliographystyle{splncs04}
\bibliography{reference}
\end{document}